\documentclass{article}

\usepackage{arxiv}

\usepackage[utf8]{inputenc} % allow utf-8 input
\usepackage[T1]{fontenc}    % use 8-bit T1 fonts
\usepackage{hyperref}       % hyperlinks
\usepackage{url}            % simple URL typesetting
\usepackage{booktabs}       % professional-quality tables
\usepackage{amsfonts}       % blackboard math symbols
\usepackage{nicefrac}       % compact symbols for 1/2, etc.
\usepackage{microtype}      % microtypography
\usepackage{lipsum}		% Can be removed after putting your text content

\usepackage[pdftex]{graphicx} % Uncomment when submitting to arXiv.
\usepackage[hang,small,bf]{caption} 
\usepackage[subrefformat=parens]{subcaption} 
\usepackage{amsmath} 
\usepackage{amssymb} 
\usepackage{bm} 

\usepackage{algorithm} 
\usepackage{algorithmic} 
\makeatletter

\renewcommand{\ALG@name}{Algorithm} 
\renewcommand{\listalgorithmname}{List of Algorithms} 
\makeatother

\renewcommand{\algorithmicrequire}{\textbf{Input:}} \renewcommand{\algorithmicensure}{\textbf{Initialize:}} 

\usepackage{comment} 
\usepackage{multirow} 
\title{Knowledge Transfer Graph for Deep Collaborative Learning} 

\date{} 					% Or removing it

\author{
Soma Minami, Tsubasa Hirakawa, Takayoshi Yamashita, Hironobu Fujiyoshi \\
Chubu University \\
1200 Matsumotocho, Kasugai, Aichi, Japan \\
\texttt{\{minami@mprg.cs, hirakawa@mprg.cs, takayoshi@isc, fujiyoshi@isc\}.chubu.ac.jp} \\
}

\begin{document} 

\maketitle

\begin{comment}
\documentclass[10pt,twocolumn,letterpaper]{article}

\usepackage[dvipdfmx]{graphicx} 
\usepackage[dvipdfmx]{color}

\usepackage{cvpr}
\usepackage{times}
\usepackage{epsfig}
\usepackage{amsmath}
\usepackage{amssymb}

\usepackage[hang,small,bf]{caption} 
\usepackage[subrefformat=parens]{subcaption} 
\captionsetup{compatibility=false} 

\usepackage{amsmath} 
\usepackage{amssymb} 
\usepackage{bm} 

\usepackage{algorithm} 
\usepackage{algorithmic} 
\makeatletter

\renewcommand{\ALG@name}{Algorithm} 
\renewcommand{\listalgorithmname}{List of Algorithms} 
\makeatother

\renewcommand{\algorithmicrequire}{\textbf{Input:}} \renewcommand{\algorithmicensure}{\textbf{Initialize:}} 

\usepackage{comment} 
\usepackage{multirow} 

\usepackage[pagebackref=true,breaklinks=true,letterpaper=true,colorlinks,bookmarks=false]{hyperref}

\def\cvprPaperID{7114}
\def\httilde{\mbox{\tt\raisebox{-.5ex}{\symbol{126}}}}

\ifcvprfinal\pagestyle{empty}\fi
\begin{document}

\title{Knowledge Transfer Graph for Deep Collaborative Learning} 

\author{First Author\\
Institution1\\
Institution1 address\\
{\tt\small firstauthor@i1.org}
\and
Second Author\\
Institution2\\
First line of institution2 address\\
{\tt\small secondauthor@i2.org}
}

\maketitle
\end{comment}

\begin{abstract}
Knowledge transfer among multiple networks using their outputs or intermediate activations have evolved through extensive manual design from a simple teacher-student approach (knowledge distillation) to a bidirectional cohort one (deep mutual learning). The key factors of such knowledge transfer involve the network size, the number of networks, the transfer direction, and the design of the loss function. However, because these factors are enormous when combined and become intricately entangled, the methods of conventional knowledge transfer have explored only limited combinations. In this paper, we propose a new graph-based approach for more flexible and diverse combinations of knowledge transfer. To achieve the knowledge transfer, we propose a novel graph representation called knowledge transfer graph that provides a unified view of the knowledge transfer and has the potential to represent diverse knowledge transfer patterns. We also propose four gate functions that are introduced into loss functions. The four gates, which control the gradient, can deliver diverse combinations of knowledge transfer. Searching the graph structure enables us to discover more effective knowledge transfer methods than a manually designed one. Experimental results on the CIFAR-10, -100, and Tiny-ImageNet datasets show that the proposed method achieved significant performance improvements and was able to find remarkable graph structures. 
\end{abstract}

\section{Introduction}
Deep neural networks have accomplished significant progress by designing their internal structure (e.g., a network's module~\cite{DenseNet,PyramidNet,ResNeXt,SENet} and architecture search~\cite{NAS,PNAS,ENAS,DARTS}). The performance of existing networks can be further improved by knowledge transfer among multiple networks, such as knowledge distillation~(KD)~\cite{KD} and deep mutual learning~(DML)~\cite{DML}, in extensive tasks without any additional dataset. These methods, which we call ``collaborative learning,'' transfer knowledge between multiple networks using their outputs and/or intermediate activations.

Collaborative learning has been manually designed in extensive studies~\cite{KD,FitNets,Born_Again_Net,TA_distillation,DML,CL,ONE,DualStudent}, including the simple teacher-student approach~\cite{KD}, self-distillation~\cite{Born_Again_Net}, an intermediation by teacher assistant~\cite{TA_distillation}, and the bidirectional cohort approach~\cite{DML}.
The key factors of such collaborative learning are the network size, the number of networks, the transfer direction, and the design of the loss function.
In general, increasing the number of networks tends to improve the performance of the target network~\cite{DML,TA_distillation,CL,ONE}.
Cho~\textit{et al.}~\cite{Efficacy_KD} also pointed out that larger models do not often make better teachers.
The methods of conventional knowledge transfer have explored only limited combinations because the combination of the key factors is enormous and has become intricately entangled. 
Therefore, it is necessary to extensively explore diverse patterns of collaborative learning to achieve more effective knowledge transfer.

\begin{figure*}[t]
\begin{center}
\includegraphics[width=0.98\linewidth]{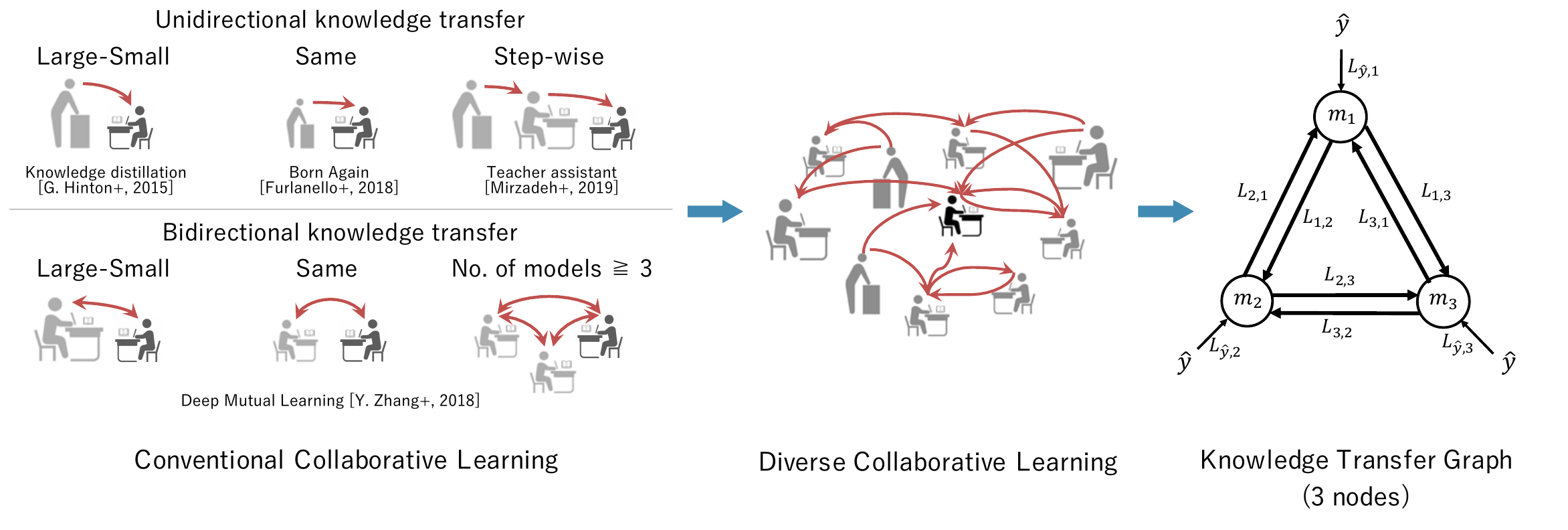}
\end{center}
%\vspace{-2zh}
\caption{\textbf{Concept of the proposed method}. From left to right, unidirectional and bidirectional knowledge transfer proposed by previous studies, the goals of our study, and the knowledge transfer graph representation we propose. In the graph, each node represents a network, each edge represents the direction of knowledge transfer, and $L_{s,t}$ represents the loss function used for training node $t$. The graph can represent diverse collaborative learning, including conventional methods.}
\label{fig:concept}
\end{figure*}

In this study, we explore more diverse knowledge transfer patterns in the above key factors for collaborative learning. Figure~\ref{fig:concept} shows the concept of our research. We propose a novel graph representation called \textit{knowledge transfer graph} that can represent both conventional and new collaborative learning. A knowledge transfer graph provides a unified view of knowledge transfer and has the potential to represent diverse knowledge transfer patterns. In the graph, each node represents a network, and each edge represents a direction of knowledge transfer. On each edge, we define a loss function that is used for transferring knowledge between the two nodes linked by the edge. Combinations of these loss functions can represent any collaborative learning with pair-wise knowledge transfer. In this paper, we propose four types of gate functions (through gate, cutoff gate, linear gate, correct gate) that are introduced into loss functions. These gate functions control the loss value, thereby delivering different effects of knowledge transfer. By arranging the loss functions at each edge, the graphs enable the representation of diverse collaborative learning patterns. Knowledge transfer graphs are searched for the network model on each node and the gate function on each edge, which enables us to discover a more effective knowledge transfer method than a manually designed one.

Our contributions are as follows. 
\begin{itemize}
\item We propose a knowledge transfer graph that represents conventional and new collaborative learning.
\item We propose four types of gates function (through gate, cutoff gate, linear gate, correct gate) to control backpropagation while training the networks. The knowledge transfer graph optimizes the gates by means of a hyperparameter search, which can achieve diverse collaborative learning.
\item We found that, compared to the vanilla model, our optimized graph achieved accuracy improvements of 1.05\% with CIFAR-10, 4.00\% with CIFAR-100, and 2.62\% with Tiny-ImageNet. On CIFAR-100, the graphs outperformed a conventional method.
\item We verified the generalization of optimized graphs that had been searched on a different dataset from the one used for optimization. In experiments using CIFAR-10, CIFAR-100, and Tiny-ImageNet, we show that the optimized graphs can be reused on another dataset.
\end{itemize}

\section{Related Work} 

\subsection{Unidirectional knowledge transfer}                   
In unidirectional knowledge transfer, the outputs of a pre-trained network are used as pseudo labels in addition to supervised labels for learning a target network effectively.
Hinton~\textit{et al.}~\cite{KD} proposed knowledge distillation, which trains a student network by using teacher network's outputs.
They succeeded in effectively transferring the teacher's internal representation to the student by introducing a temperature parameter into the softmax function.
Furlanello~\textit{et al.}~\cite{Born_Again_Net} demonstrated that KD can also train effectively in cases where the teacher network's architecture is the same as that of the student network.
Mirzadeh~\textit{et al.}~\cite{TA_distillation} proposed a method that adds a middle network, called a teacher assistant, between a teacher and student.
When there is a large performance gap between the teacher and the student, students can be effectively trained by separating them with a middle network.
Various approaches that transfer from intermediate layers have been also proposed, e.g. hint~\cite{FitNets}, flow of activations between layers~\cite{FSP_matrix}, and attention map~\cite{AT}.
\cite{RKD,metrics_from_teachers,IRG} transfer mutual relations of data samples in a mini-batch.
Distillation has been applied to object detection~\cite{detection}, domain adaptation~\cite{domain}, text-to-speech~\cite{parallel-wavenet}, etc.

\begin{comment}
Distillation has recently been connected to curriculum learning~\cite{Curriculum_learning} in information learning theory~\cite{generalized_distillation}.

Distillation has recently been connected to ~\cite{generalized_distillation}.

theory of on learning with privileged information (Pechyony & Vapnik, 2010).

pointed out some connections between distillation and curriculum learning~\cite{Curriculum_learning} in information learning theory~.
\end{comment}

\subsection{Bidirectional knowledge transfer}                   

In the bidirectional method, which was first proposed by Zhang~\textit{et al.}~\cite{DML}, there is no pre-trained teacher; randomly initialized students teach each other by transferring their knowledge.
%Even when using networks with identical structures, the accuracy is improved due to the effects of entropy regularization~\cite{Entropy-sgd}.
Even when using networks with identical structures, the accuracy is improved.
Zhang~\textit{et al.} pointed out that DML is connected to entropy regularization~\cite{Entropy-sgd, penalizing_confidence}.
In this method, all loss functions used in each network are identical.
There could be more potential variants in collaborative learning if a combination of different loss functions was used.
Further improvements in accuracy can be achieved by using the ensemble outputs of collaboratively trained networks as teachers~\cite{CL,ONE}, and by sharing the intermediate layers of these networks \cite{CL, ONE, DKS}. 
DML has been applied to large scale distributed training~\cite{parallel-SGD} and re-identification~\cite{ReID}.
Dual student~\cite{DualStudent} is a method of bidirectional knowledge transfer in semi-supervised learning.

\begin{figure*}[t]
\begin{center}
\includegraphics[width=0.95\linewidth]{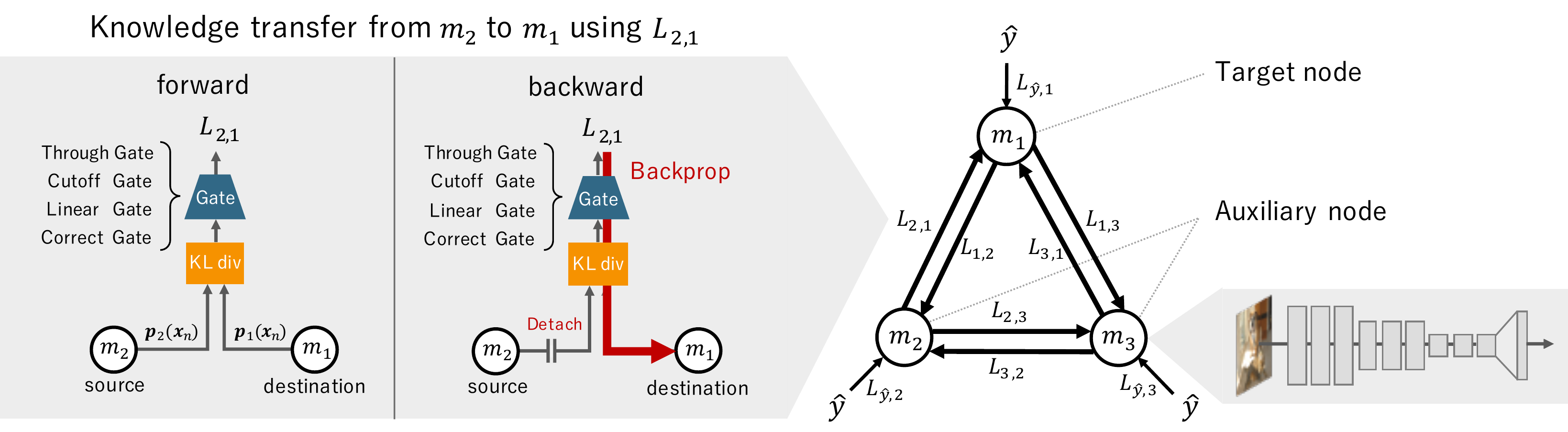}
\end{center}
\caption{\textbf{Knowledge transfer graph} (for 3-node case). Each node represents a model, and a loss function $L_{s,t}$ is defined for each edge. $\hat{y}$ is a label. $L_{s,t}$ calculates the KL divergence from the outputs of two nodes and then passes it through a gate function. The calculated loss gradient information is only propagated in the direction of the arrow. We can also represent unidirectional knowledge transfer by cutting off edges with a cutoff gate.} 
\label{fig:overview}
\end{figure*}

\section{Proposed Method}

We explore graph structures representing diverse knowledge transfer by combining loss functions with four types of gate.
We describe how to represent knowledge transfer graphs in Sec.~\ref{sec:graph_repr}, loss function of our proposed method in Sec.~\ref{sec:loss_func}, four types of gate function in Sec.~\ref{sec:gates}, optimization method of each model in Sec.~\ref{sec:network_optim}, and graph optimization method in Sec.~\ref{sec:graph_optim}.

\subsection{Knowledge transfer graph representation\label{sec:graph_repr}} 

Figure~\ref{fig:overview} shows the knowledge transfer graph representation with three nodes. 
In the proposed method, the direction of knowledge transfer between networks is represented by a directed graph, and a different loss function is defined for each edge.
By defining different loss functions, it is possible to express various knowledge transfer methods. 

We define a directed graph where node $m_i$ represents the $i$th model used for training.
Each edge represents the directions in which gradient information is transferred.
In this paper, we refer to a node that transfers its knowledge to another as a source node, and a node to which the source node transfers its knowledge as a destination node.
The losses calculated from the outputs of the two models are back-propagated towards the destination node. Losses are not back-propagated to the source node.

\subsection{Loss function\label{sec:loss_func}}
The mini-batch comprising the image of the $n$th sample $\bm{x}_n$ and the label $\hat{y}_n$ is represented as $\mathcal{B} = \{\bm{x}_n, \hat{y}_n\}_{n=1}^{N}$, and the batch size of mini-batch $\mathcal{B}$ is represented as $|\mathcal{B}|$. The label $\hat{y}_n$ represents class id. The number of models used for learning is $M$, and the source and destination nodes are $m_s$ and $m_t$, respectively. 

When obtaining the difference in output probabilities between nodes, we use the Kullback-Leibler (KL) divergence $KL(\bm{p}_{s}(\bm{x}_n) || \bm{p}_{t}(\bm{x}_n))$. Here, $\bm{p}_s$ and $\bm{p}_t$ are the outputs of the source and destination nodes, respectively, and consist of probability distributions normalized by the softmax function. 

If the one-hot vector representation of the label $\hat{y}_n$ is $\bm{p}_{\hat{y}_n}$, the loss between $\bm{p}_{\hat{y}_n}$ and the output $\bm{p}_t(\bm{x}_n)$ of destination node $t$ is calculated using the cross-entropy function $H(\bm{p}_{\hat{y}_n}, \bm{p}_t(\bm{x}_n))$. $H(\bm{p}_{\hat{y}_n}, \bm{p}_t(\bm{x}_n))$ can be decomposed into the sum of KL divergence and entropy as follows: 
\begin{equation} 
\begin{split} 
H(\bm{p}_{\hat{y}_n}, \bm{p}_t(\bm{x}_n)) &= KL(\bm{p}_{\hat{y}_n} || \bm{p}_t(\bm{x}_n)) + H(\bm{p}_{\hat{y}_n}, \bm{p}_{\hat{y}_n}) \\ &= KL(\bm{p}_{\hat{y}_n} || \bm{p}_t(\bm{x}_n)).
\end{split} 
\label{eq:CE2KL} 
\end{equation} 
Here, since $\bm{p}_{\hat{y}_n}$ is a one-hot vector, its entropy $H(\bm{p}_{\hat{y}_n}, \bm{p}_{\hat{y}_n})$ is zero.
%and so $H(\bm{p}_{\hat{y}_n}, \bm{p}_t(\bm{x}_n)) = KL(\bm{p}_{\hat{y}_n} || \bm{p}_t(\bm{x}_n))$.
Therefore, the loss between the label and the output can also be represented by the KL divergence in the same way as the loss between the node outputs. In the following, $\bm{p}_{\hat{y}_n}$ is denoted by $\bm{p}_0(\bm{x}_n)$.

$L_{s,t}$ represents the loss function used when knowledge is propagated from the source node $m_s$ to the destination node $m_t$, which is defined by
\begin{equation} 
L_{s,t} = \sum_{n}^{|\mathcal{B}|} G_{s,t}(KL(\bm{p}_{s}(\bm{x}_n) || \bm{p}_{t}(\bm{x}_n))),
\label{} 
\end{equation} 
where $G_{s,t}(\cdot)$ is a gate function.
%The gate determines which value is to be back-propagated out of the KL divergence values calculated for each sample. 

Finally, the loss function of the destination node $m_t$ is expressed as the sum of losses for all nodes as follows: 
\begin{equation} 
L_{t} = \sum_{s=0, s \neq t}^{M} L_{s,t}.
\label{eq:Loss_func} 
\end{equation} 

\subsection{Gates\label{sec:gates}} 

If all information is transferred to the destination node from the source node throughout the entire training phase, the learning of the destination node is liable to be disrupted.
We introduce a gate that controls the gradient to a destination node by weighting losses for each training sample.
%Gates weight the losses to control the propagation of gradients through the network.
We define four types of gate: through gate, cutoff gate, linear gate, and correct gate. A through gate simply passes through the losses of each training sample without any changes.

\begin{equation} 
G_{s,t}^{Through}(a) = a
\label{} 
\end{equation} 

A cutoff gate is a gate that performs no loss calculation. It can be used to cut off any edge in a knowledge transfer graph. This function is required in methods such as KD, where knowledge transfer is only performed in one direction. 
\begin{equation} 
G_{s,t}^{Cutoff}(a) = 0 
\label{} 
\end{equation} 

A linear gate changes its loss weighting linearly with time during training. It has a small weighting at the initial epoch, and its weighting becomes larger as training progresses. 
\begin{equation} 
G_{s,t}^{Linear}(a) = \frac{k}{k_{end}}a 
\label{} 
\end{equation} 
Here, $k$ is the cumulative number of updates, and $k_{end}$ is the number of updates at the end of training. 

A correct gate is a gate that only passes the losses of samples whose source node is correct. If the top-1 class number of a source node $m_s$ is $y_s$, a correct gate can be expressed as 
\begin{equation} 
G_{s,t}^{Correct}(a; \hat{y}, y_s) = \delta_{\hat{y}, y_s} \cdot a.
\label{} 
\end{equation} 
When the source node is not a pre-trained model, the propagation of false information can be suppressed at the initial epoch. While a linear gate weights the overall loss, a correct gate selects the samples from which the loss is calculated.

\subsection{Network optimization\label{sec:network_optim}} 

Algorithm~\ref{alg1} shows how to update the network of each node during training. First, all the model weights are randomly initialized unless all the gates $G_{i,t}$ corresponding to nodes $m_i$ are cutoff gates, in which case $m_i$ is initialized with the weights of the pre-trained model. The pre-trained model is trained only with the labels, using the same dataset as the one used for the hyperparameter search. Here, $m_i$ is frozen during training and its weights are not updated. This node performs a role being equivalent to that of the teacher network used in KD. 

The losses are obtained by inputting the same samples to all nodes. Gradients are obtained from the resulting losses, and all nodes are updated simultaneously. The gradient of loss $L_t$ obtained from Eq.~(\ref{eq:Loss_func}) is back-propagated only to node $m_t$, and has no effect on the other nodes. In DML, after updating the weights of the first node, the training samples are input again to the updated nodes to obtain an output. The losses between every node are then recalculated from this outputs, and gradient descent is performed for the second node. These steps are repeated until every node has been updated. The drawback of DML is that this updating method causes a significant increase in computational cost as the number of nodes increases. In our proposed method, since the weights of every node are updated during a single forward calculation, it is possible to reduce the computational cost during training.

\begin{algorithm}[t] 
\caption{Network Optimization} 
\label{alg1} 
\begin{algorithmic}[] 
\REQUIRE Number of nodes $M$, number of epochs $E$
\ENSURE Initialize all network weights, or read in the weights of a pre-trained network 
\FOR{\_ = 1 to E} 
	\STATE Input the same image to each network $m_n$, and obtain the output $\bm{p}_n$. 
	\STATE Obtain the loss $L_{n}$ according to Eq.~(\ref{eq:Loss_func}). 
	\STATE Obtain the update quantity of $m_n$ from the gradient $L_{n}$. 
	\STATE Update the weights of all networks. 
\ENDFOR
\end{algorithmic} 
\end{algorithm}

\subsection{Graph optimization\label{sec:graph_optim}}
We refer to an optimized node by hyperparameter search as a target node $m_1$, and a node that supports training of the target node as an auxiliary node. A target node to be optimized is specified, and the knowledge transfer graph is optimized to maximize the accuracy of this node. The hyperparameters to be optimized are the model type of the auxiliary node and the gate type on each edge. The size of the search space for this optimization is $M^{(n-1)} \cdot G^{N^2} / 2$, where $N$ is the number of nodes, $M$ is the number of model types, and $G$ is the number of gate types. For example, if $N = 3$, $M = 3$, and $G = 4$, there are over one million patterns.

We used the Asynchronous Successive Halving Algorithm~(ASHA)~\cite{ASHA} as the hyperparameter optimization method. First, using $D$ GPU servers, we randomly create a knowledge transfer graph with $D$ servers and perform distributed asynchronous learning. In each knowledge transfer graph, the accuracy of the target node is evaluated using verification data at epochs $1, 2, 4, \cdots, 2^k$. If this accuracy is in the lower 50\% of all the accuracy values evaluated in the past, the graph is abandoned and training is performed again after generating a new graph. This process is repeated until the total number of trials reaches $T$. ASHA can achieve improvements in terms of both temporal efficiency and accuracy by performing a random search with active early termination in a parallel distributed environment. We performed optimization with $D=30$ and $T=1500$. 

\section{Experiments} 
We performed experiments to determine the efficacy of knowledge transfer graphs searched by ASHA. 

\subsection{Experimental setting} 
\textbf{Dataset:} We used the CIFAR-10, CIFAR-100~\cite{Cifar}, and Tiny-ImageNet~\cite{Tiny-ImageNet} datasets, which are typically used for general object recognition. CIFAR-10 and CIFAR-100 consist of 50,000 images for training and 10,000 images for verification. Both datasets consist of images with dimensions of 32$\times$32 pixels and include labels for 10 and 100 classes, respectively. Data augmentation was performed by processing the training images with 4-pixel padding (reflection), random cropping, and random flipping. Data augmentation was not applied to the verification images. The Tiny-ImageNet dataset consists of 100,000 training images and 10,000 verification images sampled from the ImageNet \cite{ILSVRC15} dataset. This dataset consists of images with dimensions of 64$\times$64 pixels and labels for 200 classes. The data augmentation settings were the same as those for the CIFAR datasets. 

\textbf{Models:} We used three networks: ResNet32, ResNet110~\cite{ResNet}, and Wide ResNet 28-2~\cite{WRN}. Table~\ref{table:independent_accuracy} shows the accuracy achieved when each model was trained with supervised labels only. However, when training with Tiny-ImageNet, since the images are larger in size, the stride of the initial convolution layer was set to 1.

\textbf{Implementation details:} For the optimization algorithm, we used SGD and Nesterov momentum in all experiments. The initial learning rate was 0.1, the momentum was 0.9, and the batch size was 64. When training on CIFAR, the learning rate was reduced to one tenth every 60 epochs, for a total of 200 epochs. When training on the Tiny-ImageNet, the learning rate was reduced to one tenth at the 40th, 60th, and 70th epochs, for a total of 80 epochs. The reported accuracy values are averaged over five trials with a fixed graph structure implemented after obtaining the optimized graph. The standard deviation over each set of five trials is also shown. Our experiments were implemented using the Pytorch framework for deep learning and the Optuna framework for hyperparameter searching. The computations were performed using 90 Quadro P5000 servers. Our implementation is available at \{the URL of our github code: It will be available at camera ready.\} 

\begin{figure*}[t]
%\vspace{-0zh}
\begin{tabular}{ccc}
\begin{minipage}[t]{0.24\hsize}
\centering
\includegraphics[width=1.1\linewidth]{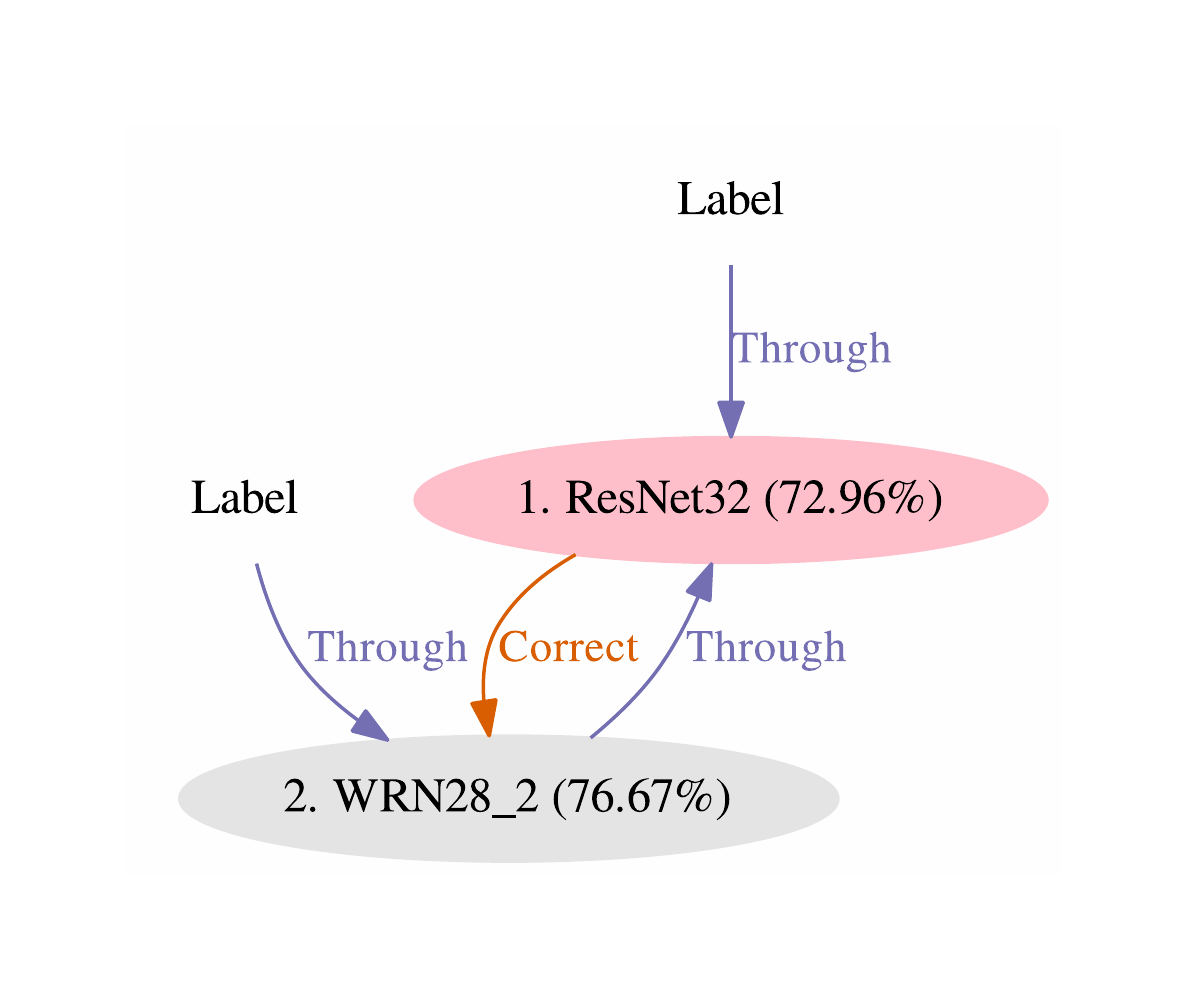}
\subcaption{\textbf{2 nodes (72.96\%)}}
\label{fig:02models}
\end{minipage} &
\begin{minipage}[t]{0.26\hsize}
\centering
\includegraphics[width=1.3\linewidth]{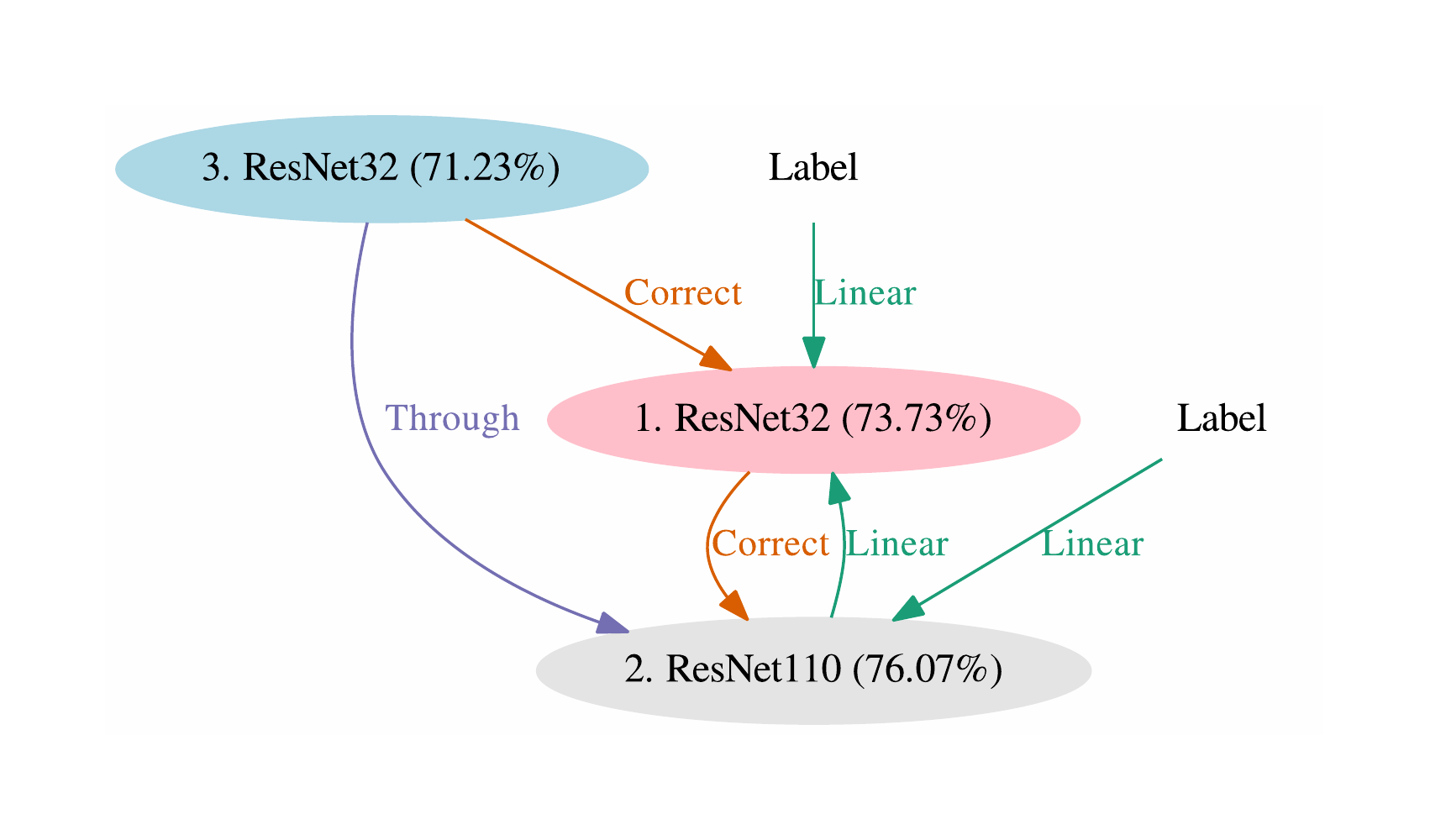}
\subcaption{\textbf{3 nodes (73.73\%)}}
\label{fig:03models}
\end{minipage} &
\begin{minipage}[t]{0.40\hsize}
\centering
\includegraphics[width=1.0\linewidth]{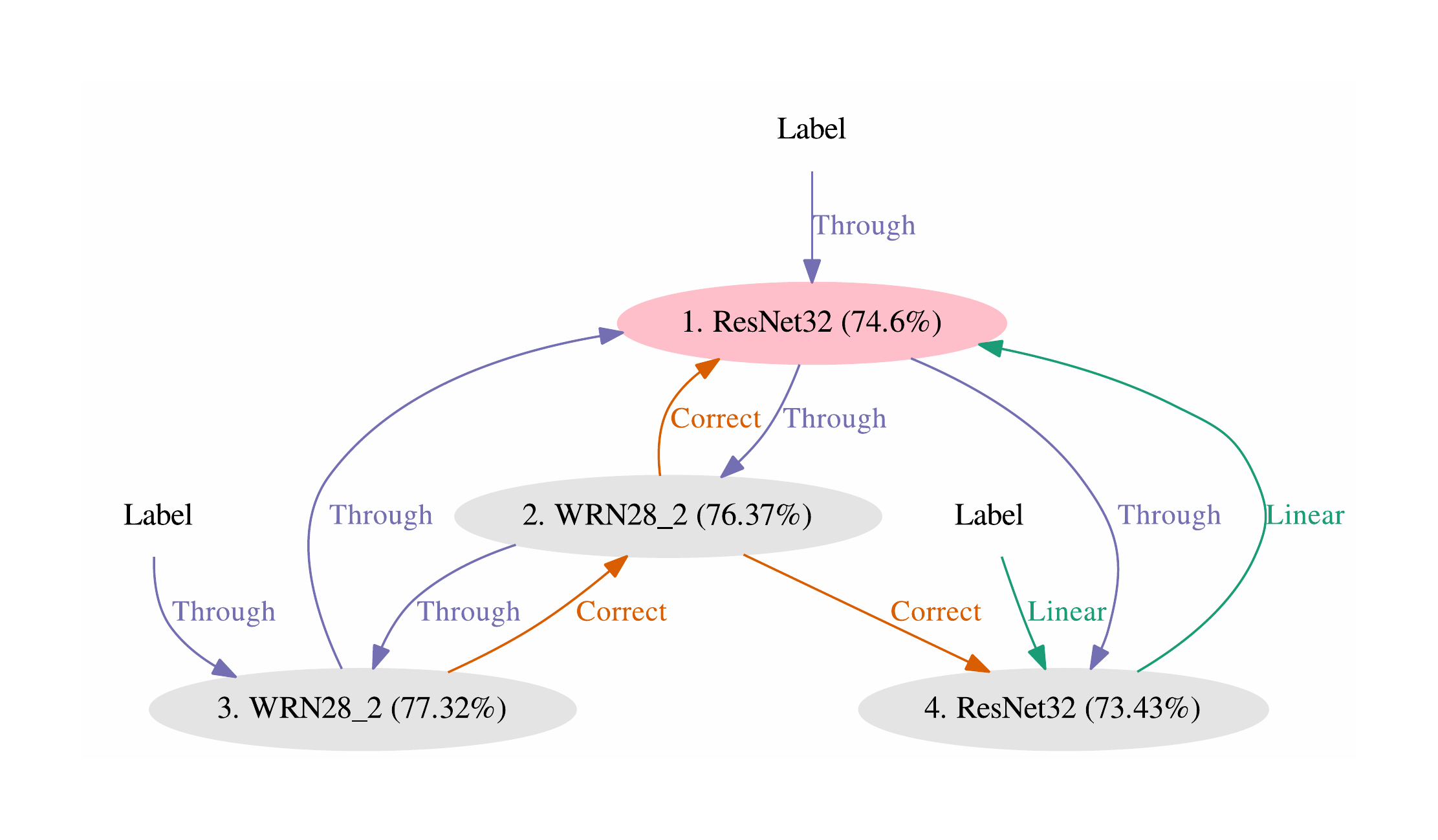}
\subcaption{\textbf{4 nodes (74.60\%)}}
\label{fig:04models}
\end{minipage} \\
\begin{minipage}[t]{0.24\hsize}
\centering
\includegraphics[width=1.1\linewidth]{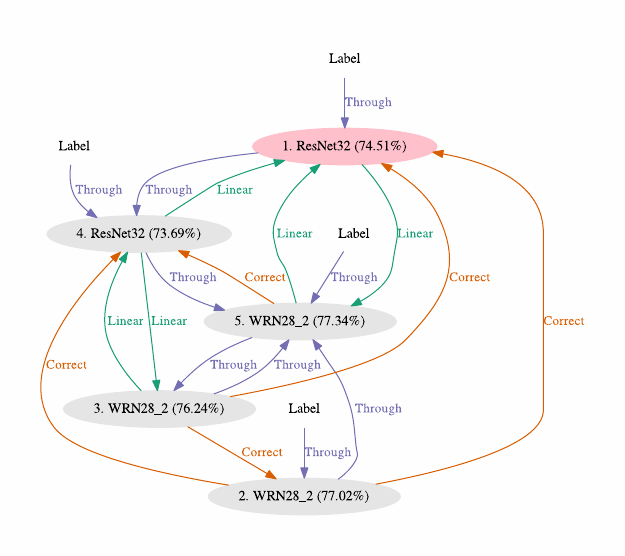}
\subcaption{\textbf{5 nodes (74.51\%)}}
\label{fig:05models}
\end{minipage} &
\begin{minipage}[t]{0.28\hsize}
\centering
\includegraphics[width=1.15\linewidth]{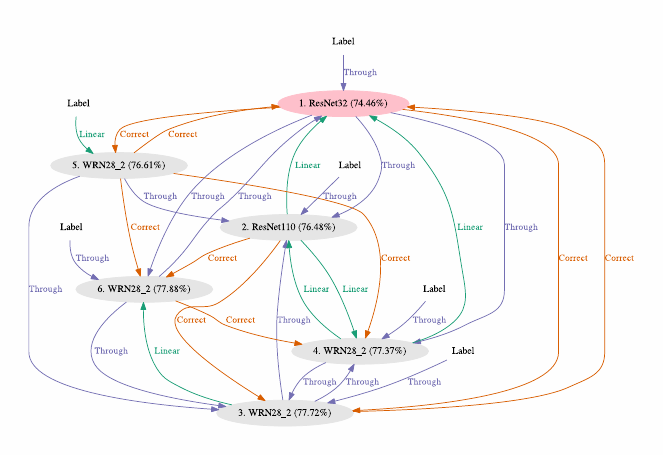}
\subcaption{\textbf{6 nodes (74.46\%)}}
\label{fig:06models}
\end{minipage} &
\begin{minipage}[t]{0.40\hsize}
\centering
\includegraphics[width=1.0\linewidth]{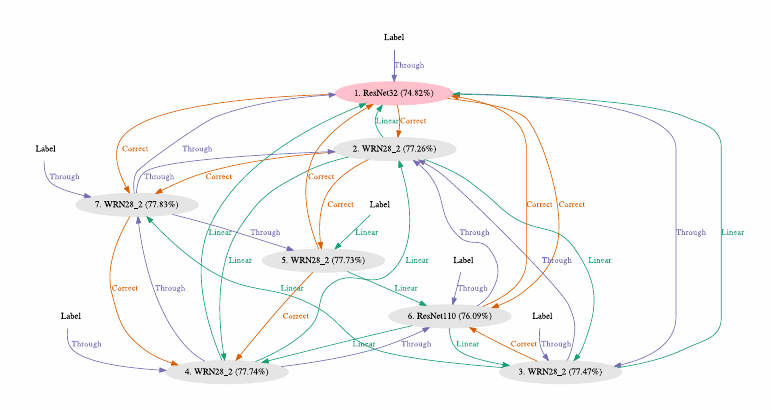}
\subcaption{\textbf{7 nodes (74.82\%)}}
\label{fig:07models}
\end{minipage}

\end{tabular}
\caption{\textbf{Optimized knowledge transfer graph}. Red node is the target node, blue node is the pre-trained node, and ``Label'' represents supervised labels. At each edge, the selected gate is shown, exclusive of cutoff gate. Numbers in parentheses show the accuracy achieved in one out of five trials.}
\label{fig:graph}
\end{figure*}

\begin{table}[t]
\begin{center}
\begin{tabular}{c|c}
Model & Accuracy [\%] \\ \hline
ResNet32 & 70.71 $\pm$ 0.39 \\
ResNet110 & 72.59 $\pm$ 0.54 \\
Wide ResNet 28-2 & 74.60 $\pm$ 0.38 
\end{tabular}
\end{center}
\caption{\textbf{Accuracy of vanilla models}. Mean and standard deviation of single network accuracies on test data.}
\label{table:independent_accuracy}
\end{table}

\begin{table*}[t]
\begin{center}
\begin{tabular}{c|c|cccc|cc}
Method & Accuracy (Node 1) & Node 1 & Node 2 & Node 3 & Node 4 & TeCost & TrCost \\ \hline \hline
Vanilla & 70.71 $\pm$ 0.39 & ResNet32 & – & – & – & 70.2 & 70.2 \\ \hline
DML~\cite{DML} & 72.00 $\pm$ 0.44 & ResNet32 & ResNet32 & – & – & 70.2 & 140.5 \\
KD ($T=2$)~\cite{KD} & 71.88 $\pm$ 0.78 & ResNet32 & WRN28-2* & – & – & 70.2 & 286.8 \\
DML~\cite{DML} & 72.71 $\pm$ 0.18 & ResNet32 & WRN28-2 & – & – & 70.2 & 286.8 \\
Ours & \textbf{72.88} $\pm$ 0.41 & ResNet32 & WRN28-2 & – & – & 70.2 & 286.8 \\ \hline
DML~\cite{DML} & 72.09 $\pm$ 0.43 & ResNet32 & ResNet32 & ResNet32 & – & 70.2 & 210.8 \\
DML~\cite{DML} & 72.20 $\pm$ 0.47 & ResNet32 & ResNet110 & ResNet32 & – & 70.2 & 398.2 \\
Ours & \textbf{73.46} $\pm$ 0.42 & ResNet32 & ResNet110 & ResNet32* & – & 70.2 & 398.2 \\ \hline
DML~\cite{DML} & 72.76 $\pm$ 0.35 & ResNet32 & ResNet32 & ResNet32 & ResNet32 & 70.2 & 281.0 \\
Song~\textit{et al.}~\cite{CL} & 73.68** $\pm$ 0.26 & \multicolumn{4}{|c|}{(4$\times$ResNet32 with shared intermediate layers)} & 70.2 & 160.9 \\
ONE~\cite{ONE} & 73.42** $\pm$ N/A & \multicolumn{4}{|c|}{(4$\times$ResNet32 with shared intermediate layers)} & 70.2 & 138.1 \\
DML~\cite{DML} & 73.32 $\pm$ 0.55 & ResNet32 & WRN28-2 & WRN28-2 & ResNet32 & 70.2 & 573.6 \\
Ours & \textbf{74.34} $\pm$ 0.32 & ResNet32 & WRN28-2 & WRN28-2 & ResNet32 & 70.2 & 573.6
\end{tabular}
\caption{\textbf{Comparison with conventional methods}. ``*'' denotes a pre-trained model. $T$ is a temperature parameter. ``**'' denotes a value cited from the paper. TeCost and TrCost represent the computational complexity (MFLOPs) for testing and training, respectively.}
\label{table:vscomparison}
\end{center}
\end{table*}

\begin{table*}[t] 
\begin{center} 

\begin{tabular}{cc|ccc} No. of nodes &Gates &CIFAR-10 &CIFAR-100 &Tiny-ImageNet \\ \hline \hline
1 &– &93.12 $\pm$ 0.27 &70.71 $\pm$ 0.39 &53.18 $\pm$ 0.08\\ \hline
\multirow{2}{*}{2} &Fixed (Through) &93.25 $\pm$ 0.50 &72.47 $\pm$ 0.38 & 54.93 $\pm$ 0.29 \\
 &Optimized &93.65 $\pm$ 0.14 & 72.88 $\pm$ 0.41 &54.69 $\pm$ 0.16 \\ \hline
\multirow{2}{*}{3} &Fixed (Through) &93.53 $\pm$ 0.24 &71.88 $\pm$ 0.43 &53.78 $\pm$ 0.78 \\
 &Optimized &93.92 $\pm$ 0.20 & 73.46 $\pm$ 0.42 & 55.02 $\pm$ 0.31 \\ \hline
\multirow{2}{*}{4} &Fixed (Through) &93.01 $\pm$ 0.79 &73.40 $\pm$ 0.39 &53.92 $\pm$ 0.21 \\
 &Optimized &93.99 $\pm$ 0.27 & 74.34 $\pm$ 0.32 &\textbf{55.80} $\pm$ 0.26 \\ \hline
\multirow{2}{*}{5} &Fixed (Through) &93.61 $\pm$ 0.23 &73.40 $\pm$ 0.28 &52.12 $\pm$ 0.30 \\
 &Optimized &94.14 $\pm$ 0.16 & 74.54 $\pm$ 0.59 & 55.30 $\pm$ 0.16 \\ \hline
\multirow{2}{*}{6} & Fixed (Through) & 93.84 $\pm$ 0.39 & 73.85 $\pm$ 0.45 & 49.37 $\pm$ 1.70 \\
& Optimized & \textbf{94.17} $\pm$ 0.21 & 74.22 $\pm$ 0.22 & 55.16 $\pm$ 0.19 \\ \hline
\multirow{2}{*}{7} & Fixed (Through) & 93.75 $\pm$ 0.27 & 73.53 $\pm$ 0.27 & 53.10 $\pm$ 0.44 \\
& Optimized & 94.07 $\pm$ 0.14 & \textbf{74.71} $\pm$ 0.23 & 54.78 $\pm$ 0.36
\end{tabular} 
\end{center} 
\caption{\textbf{Results of optimization on various datasets}. ResNet32 was used as the target node. ``Fixed'' indicates all gates are through gates. ``Optimized'' indicates they have been optimized. } 
\label{table:various_dataset} 
\end{table*} 

\begin{table*}[t]
\centering
\begin{tabular}{c|ccc||ccc}
& \multicolumn{3}{c||}{evaluated on CIFAR-100} & \multicolumn{3}{c}{evaluated on Tiny-ImageNet} \\ \cline{2-7} 
No. of nodes & \begin{tabular}[c]{@{}c@{}}fixed to\\ through gate\end{tabular} & \begin{tabular}[c]{@{}c@{}}searched on\\ different dataset\\ (CIFAR-10)\end{tabular} & \begin{tabular}[c]{@{}c@{}}searched on\\ same dataset\\ (CIFAR-100)\end{tabular} & \begin{tabular}[c]{@{}c@{}}fixed to\\ through gate\end{tabular} & \begin{tabular}[c]{@{}c@{}}searched on\\ different dataset\\ (CIFAR-10)\end{tabular} & \begin{tabular}[c]{@{}c@{}}searched on\\ same dataset\\ (Tiny-ImageNet)\end{tabular} \\ \hline \hline
2 & 72.47 $\pm$ 0.38 & 72.50 $\pm$ 0.33 & \textbf{72.88} $\pm$ 0.41 & \textbf{54.93} $\pm$ 0.29 & 52.79 $\pm$ 0.31 & 54.69 $\pm$ 0.16 \\
3 & 71.88 $\pm$ 0.43 & \textbf{73.63} $\pm$ 0.18 & 73.46 $\pm$ 0.42 & 53.78 $\pm$ 0.78 & 54.21 $\pm$ 0.44 & \textbf{55.02} $\pm$ 0.31 \\
4 & 73.40 $\pm$ 0.39 & 73.76 $\pm$ 0.25 & \textbf{74.34} $\pm$ 0.32 & 53.92 $\pm$ 0.21 & 54.50 $\pm$ 0.36 & \textbf{55.80} $\pm$ 0.26 \\
5 & 73.40 $\pm$ 0.28 & \textbf{74.62} $\pm$ 0.24 & 74.54 $\pm$ 0.59 & 52.12 $\pm$ 0.30 & 54.42 $\pm$ 0.14 & \textbf{55.30} $\pm$ 0.16 
\end{tabular}
\caption{\textbf{Accuracy rate of reused graphs optimized on another dataset}. Graphs are trained on CIFAR-100 and Tiny-ImageNet, where graphs are searched on CIFAR-10. Target node is ResNet32. }
\label{table:10to100}
\end{table*}

\subsection{Optimized knowledge transfer graphs \label{Sec:graph_visualization}}
Figure~\ref{fig:graph} shows the visualization of the knowledge transfer graphs with two to seven nodes optimized on CIFAR-100. The graph for two nodes (Fig.~\ref{fig:02models}) is similar to the conventional bidirectional method, where only the edge from ResNet32 to Wide ResNet 28-2 is the correct gate, and the other edges are the through gate. From the small network (ResNet32) to the large network (Wide ResNet), only the correct outputs that match the supervised labels are transferred. The graph for three nodes (Fig.~\ref{fig:03models}) is a fusion of unidirectional and bidirectional knowledge transfer. At the beginning of training, the target node (node 1) imitates only the pre-trained node 3, which is a unidirectional method, because the linear gate and correct gate deactivate the specific edges between the target node and the labels and between the target node and node 2. As training progresses, linear and correct gates activate those edges gradually, resulting in unidirectional and bidirectional knowledge transfer. In the graph for four nodes (Fig.~\ref{fig:04models}), two of them are ResNet32 and the other two are Wide ResNet28-2. Knowledge of supervised labels is not directly transferred to node 2, and the node acts as an intermediary for other nodes, like a teacher assistant~\cite{TA_distillation}.

For all numbers of nodes, the target node had much better accuracy than that in individual learning (see Tab.~\ref{table:independent_accuracy}). The accuracy of nodes other than the target node was also improved. We found that ResNet32 and ResNet110 were selected as the nodes of top-1 graphs as well as the highest performance Wide ResNet 28-2, and the performance of the target node tended to improve when the number of nodes was increased. Our quantitative evaluation is discussed in Sec.~\ref{sec:N-nodes_graghs}. 

\subsection{Comparison with conventional methods}
Table \ref{table:vscomparison} compares the results of the proposed and conventional methods. ``Ours'' shows the results of the proposed method for optimized graphs with two, three, or four nodes. ``KD~\cite{KD}'' uses a pre-trained Wide ResNet 28-2 network as a teacher, and sets the temperature parameter to $T=2$. In ``DML~\cite{DML}'' using over three nodes, all student networks have the same architecture. Since the proposed method chooses which model to use as a hyper parameter, it is possible to select the optimal combination of models. In ``Song~\textit{et al.}~\cite{CL}'' and ``ONE~\cite{ONE}'', the intermediate layers of multiple networks are shared during training. Then, only layers that are close to the output layer are branched, and the ensemble output of the branched output layers is used as a teacher. 

The model learned by the optimized knowledge transfer graph achieved better accuracy than KD, DML, and the latest proposed method.

\subsection{Comparison with graphs lacking diversity\label{sec:N-nodes_graghs}}
Table~\ref{table:various_dataset} shows the accuracy of target nodes in graphs searched on CIFAR-10, CIFAR-100, and Tiny-ImageNet. The comparison is a non-diverse graph, where each edge has only a through gate and each node is the same model as that of the graph, which is similar to the conventional unidirectional method~\cite{DML}.

The proposed method achieved higher accuracy than the comparative method in almost every condition, thus demonstrating the importance of using gates to control the gradient. Moreover, the optimized graphs tended to improve the accuracy when the number of nodes was increased in all datasets. The fixed gates method has the same loss function on all the edges, making it difficult to generate diversity even when the number of nodes is increased.

In our experiments, due to the limitation of computational resources, we ran only 1,500 trials for searching the knowledge transfer graphs. This may not be sufficient because the search space exponentially increases with the number of nodes. Moreover, if we searched on a larger number of trials, it might be possible to acquire a better knowledge transfer graph than we discovered. We will explore this possibility in future work.

\begin{figure*}[t]
\begin{tabular}{ccc}
\begin{minipage}[t]{0.32\hsize}
\centering
\includegraphics[width=1.1\linewidth]{03models_each.pdf}
\subcaption{\textbf{ResNet32}}
\label{fig:ResNet32}
\end{minipage} &
\begin{minipage}[t]{0.32\hsize}
\centering
\includegraphics[width=0.8\linewidth]{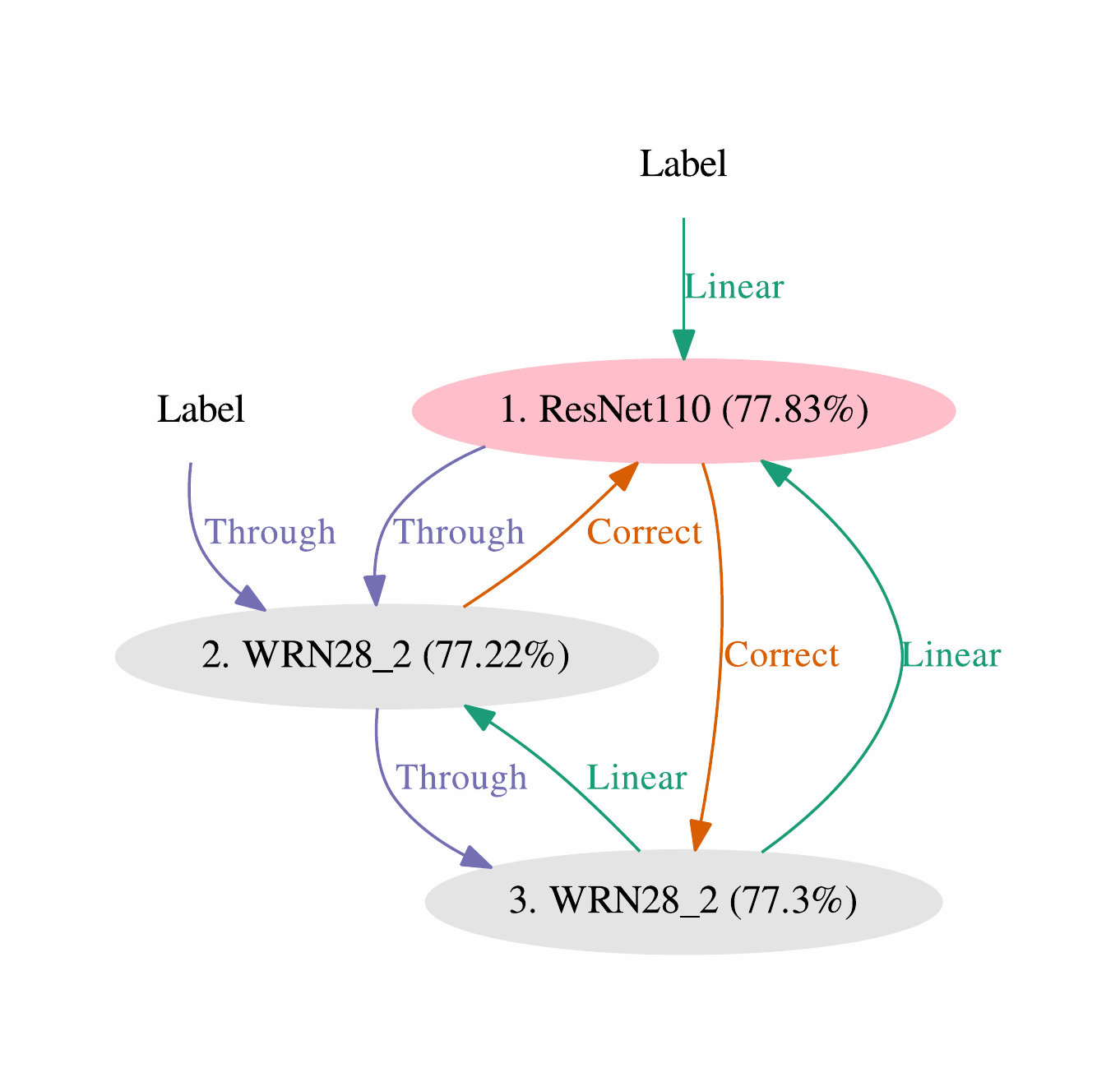}
\subcaption{\textbf{ResNet110}}
\label{fig:ResNet110}
\end{minipage} &
\begin{minipage}[t]{0.32\hsize}
\centering
\includegraphics[width=0.95\linewidth]{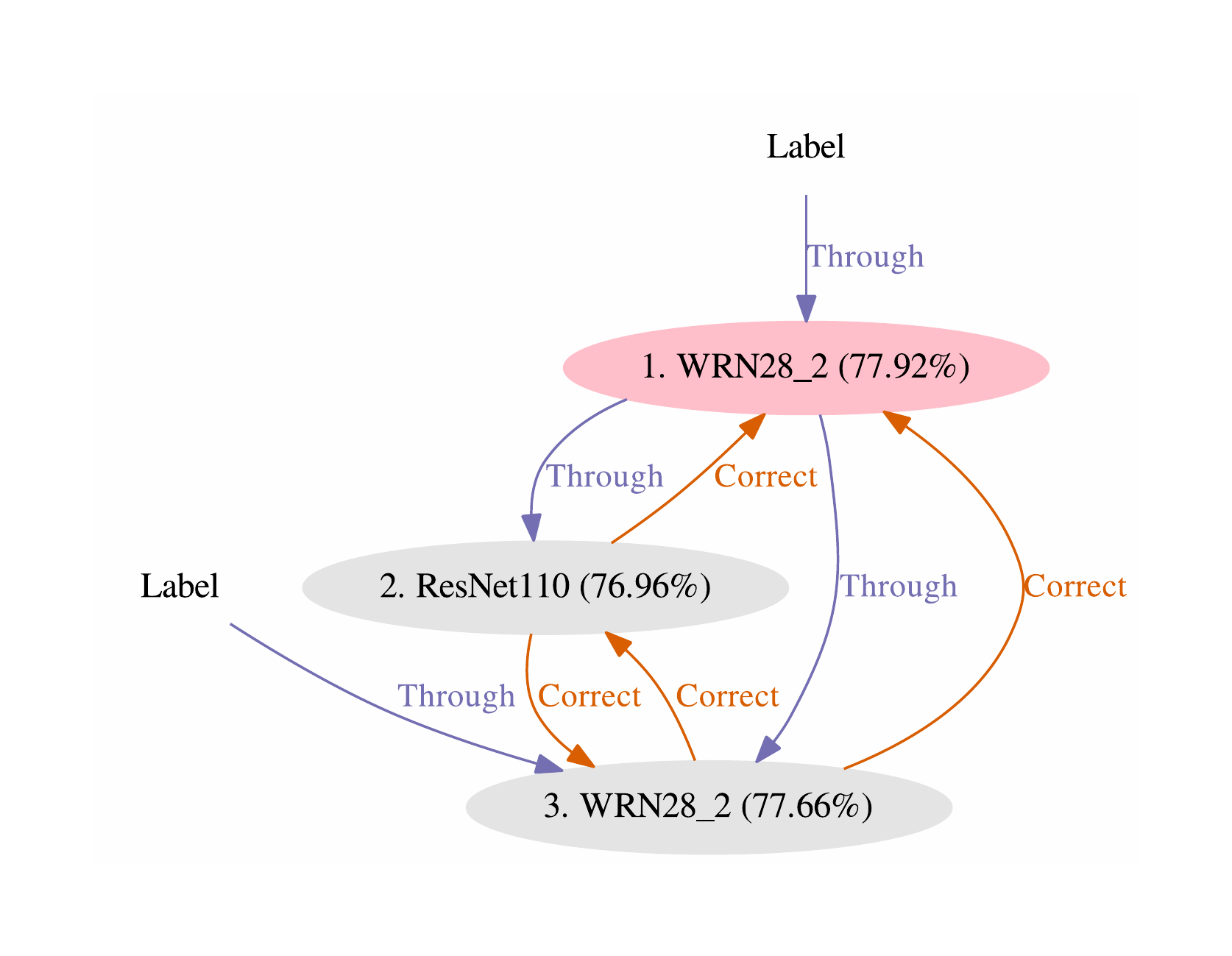}
\subcaption{\textbf{WRN28-2}}
\label{fig:WRN28-2}
\end{minipage}
\end{tabular}
\caption{\textbf{Visualization of optimized graphs with three types of target node.} Graphs whose target node is ResNet32, ResNet110, or Wide ResNet 28-2 are searched on CIFAR-100.}
\label{fig:various_target_model}
\end{figure*}

\subsection{Graph transfer to different dataset}
We investigated the generalization of graphs on different datasets. Table~\ref{table:10to100} shows the accuracy of graphs trained on CIFAR-100 and Tiny-ImageNet, where the graphs are searched on CIFAR-10. CIFAR-10, which is a 10-class dataset consisting of images of vehicles and animals, has a different distribution from CIFAR-100, which is a 100-class dataset featuring plants, insects, furniture, etc., and from Tiny-ImageNet.

When evaluated on CIFAR-100, the graphs searched on CIFAR-10 achieved the same performance as those searched on CIFAR-100. When evaluated on Tiny-ImageNet, while the performance of the graphs searched on CIFAR-10 was lower than that of graphs searched on Tiny-ImageNet, it was higher than that of non-diverse graphs whose gates were all through gates, which is a conventional bidirectional method. These results indicate that the knowledge transfer graph can be reused not only for datasets with relatively close distribution, such as CIFAR-100, but also for those with very different images, such as Tiny-ImageNet. As such, the reused graphs can greatly reduce the computational cost, since the searching process can be omitted.

\begin{table}[t]
\begin{center} 
\begin{tabular}{cc|c} Target node & Gates & Accuracy [\%] \\ \hline
\multirow{2}{*}{ResNet32} & Fixed (Through) &71.88 $\pm$ 0.43 \\ 
& Optimized &\textbf{73.46} $\pm$ 0.42 \\ \hline
\multirow{2}{*}{ResNet110} & Fixed (Through) &76.08 $\pm$ 0.92 \\ 
& Optimized &\textbf{77.63} $\pm$ 0.22 \\ \hline
\multirow{2}{*}{WRN28-2} & Fixed (Through) &77.16 $\pm$ 0.10 \\ 
& Optimized &\textbf{77.62} $\pm$ 0.24 \\
\end{tabular}
\end{center}
\caption{\textbf{Accuracy of three types of target node.} Graphs were optimized with different target nodes on CIFAR-100.}
\label{table:various_target_model}
\end{table}

\subsection{Optimization of different target nodes}
Table~\ref{table:various_target_model} shows the performances when ResNet32, ResNet110, or Wide ResNet 28-2 were set as the target node. The knowledge transfer graphs are trained on CIFAR-100 with three nodes. All the optimized graphs achieved a higher performance than the non-diverse graph, each of whose edges had only a through gate.

Figure~\ref{fig:various_target_model} shows a visualization of the optimized graphs. We found that the three nodes in the graphs included at least one model that was different from the target node. Even when Wide ResNet 28-2, which had the highest performance, was selected as the target node, the graph included ResNet110, which had a lower performance than Wide ResNet. This result suggests that diverse model selection is necessary to create a good knowledge transfer graph. To verify this hypothesis, we investigated which model had been selected for each auxiliary node of the high performance graphs. Table~\ref{table:auxiliary_node} shows the percentages of model type contained in the auxiliary nodes of the top-10 graphs for the experiment shown in Tab.~\ref{table:various_target_model}. These percentages show the same tendency among all evaluated target nodes. This result demonstrates that diverse model selection contributes to a high performance for collaborative learning.

\section{Conclusion and Future Work} 
In this paper, we propose a new learning method for more flexible and diverse combinations of knowledge transfer using a novel graph representation called knowledge transfer graph.
The graph provides a unified view of the knowledge transfer and has the potential to represent diverse knowledge transfer patterns.
We also propose four gate functions that can deliver diverse combinations of knowledge transfer.
%Searching the graph structure enables to discover more effective knowledge transfer methods than a manually designed one.
Searching the graph structure, we discovered remarkable graphs that achieved significant performance improvements.
We searched graphs over 1,500 trials, but the actual search space is much larger.
A more exhaustive search will be the focus of future work.

Since our proposed method defines nodes as individual networks, it only transfers knowledge from the output layers of these networks. Future work will include knowledge transfer from an intermediate layer. It should also be possible to perform knowledge transfer using the ensemble inference of multiple networks. Other interesting possibilities include the introduction of an encoder/decoder model, and the use of multitasking.

\begin{table}[t]
\begin{center}
\begin{tabular}{c|ccc}
& \multicolumn{3}{c}{Auxiliary node} \\
Target node & ResNet32 & ResNet110 & WRN28-2 \\ \hline
ResNet32 & 25 \% & 10 \% & 65 \% \\ \hline
ResNet110 & 15 \% & 10 \% & 75 \% \\ \hline
WRN28-2 & 30 \% & 20 \% & 50 \% \\
\end{tabular}
\end{center}
\caption{\textbf{Tendency of the selected auxiliary nodes.} Percentages of model type contained in the auxiliary nodes of the top-10 graphs in the experiment of Tab.~\ref{table:various_target_model} are shown.}
\label{table:auxiliary_node}
\end{table}

%\bibliographystyle{plain}
%\bibliography{egbib}

\begin{thebibliography}{99}

\bibitem{Tiny-ImageNet}
{Tiny ImageNet Visual Recognition Challenge}.
\newblock \url{https://tiny-imagenet.herokuapp.com/}.

\bibitem{parallel-SGD}
Rohan Anil, Gabriel Pereyra, Alexandre Passos, Robert Ormandi, George~E Dahl,
  and Geoffrey~E Hinton.
\newblock Large scale distributed neural network training through online
  distillation.
\newblock {\em arXiv preprint arXiv:1804.03235}, 2018.

\bibitem{Entropy-sgd}
Pratik Chaudhari, Anna Choromanska, Stefano Soatto, Yann LeCun, Carlo Baldassi,
  Christian Borgs, Jennifer Chayes, Levent Sagun, and Riccardo Zecchina.
\newblock Entropy-sgd: Biasing gradient descent into wide valleys.
\newblock {\em International Conference on Learning Representations (ICLR)},
  2017.

\bibitem{detection}
Guobin Chen, Wongun Choi, Xiang Yu, Tony Han, and Manmohan Chandraker.
\newblock Learning efficient object detection models with knowledge
  distillation.
\newblock In {\em Advances in Neural Information Processing Systems (NeurIPS)},
  pages 742--751, 2017.

\bibitem{domain}
Yuhua Chen, Wen Li, and Luc Van~Gool.
\newblock Road: Reality oriented adaptation for semantic segmentation of urban
  scenes.
\newblock In {\em IEEE Conference on Computer Vision and Pattern Recognition
  (CVPR)}, pages 7892--7901, 2018.

\bibitem{Efficacy_KD}
Jang-Hyun Cho and Bharath Hariharan.
\newblock On the efficacy of knowledge distillation.
\newblock In {\em International Conference on Computer Vision (ICCV)}, 2019.

\bibitem{Born_Again_Net}
Tommaso Furlanello, Zachary Lipton, Michael Tschannen, Laurent Itti, and Anima
  Anandkumar.
\newblock Born again neural networks.
\newblock In {\em International Conference on Machine Learning (ICML)},
  volume~80 of {\em Proceedings of Machine Learning Research}, pages
  1607--1616, Stockholmsm{\"a}ssan, Stockholm Sweden, 10--15 Jul 2018. PMLR.

\bibitem{PyramidNet}
Dongyoon Han, Jiwhan Kim, and Junmo Kim.
\newblock Deep pyramidal residual networks.
\newblock {\em IEEE Conference on Computer Vision and Pattern Recognition
  (CVPR)}, 2017.

\bibitem{ResNet}
Kaiming He, Xiangyu Zhang, Shaoqing Ren, and Jian Sun.
\newblock Deep residual learning for image recognition.
\newblock In {\em IEEE conference on computer vision and pattern recognition
  (CVPR)}, pages 770--778, 2016.

\bibitem{KD}
Geoffrey Hinton, Oriol Vinyals, and Jeff Dean.
\newblock Distilling the knowledge in a neural network.
\newblock {\em NIPS Deep Learning and Representation Learning Workshop}, 2015.

\bibitem{SENet}
Jie Hu, Li~Shen, and Gang Sun.
\newblock Squeeze-and-excitation networks.
\newblock 2018.

\bibitem{DenseNet}
Gao Huang, Zhuang Liu, Laurens Van Der~Maaten, and Kilian~Q Weinberger.
\newblock Densely connected convolutional networks.
\newblock In {\em IEEE Conference on Computer Vision and Pattern Recognition
  (CVPR)}, volume~1, page~3, 2017.

\bibitem{DualStudent}
Zhanghan Ke, Daoye Wang, Qiong Yan, Jimmy Ren, and Rynson~W.H. Lau.
\newblock Dual student: Breaking the limits of the teacher in semi-supervised
  learning.
\newblock In {\em Proceedings of the IEEE International Conference on Computer
  Vision}, 2019.

\bibitem{Cifar}
Alex Krizhevsky and Geoffrey Hinton.
\newblock Learning multiple layers of features from tiny images.
\newblock Technical report, Citeseer, 2009.

\bibitem{ONE}
Xu~Lan, Xiatian Zhu, and Shaogang Gong.
\newblock Knowledge distillation by on-the-fly native ensemble.
\newblock In {\em Advances in Neural Information Processing Systems (NeurIPS)},
  pages 7527--7537, 2018.

\bibitem{ASHA}
Liam Li, Kevin Jamieson, Afshin Rostamizadeh, Ekaterina Gonina, Moritz Hardt,
  Benjamin Recht, and Ameet Talwalkar.
\newblock Massively parallel hyperparameter tuning.
\newblock {\em arXiv preprint arXiv:1810.05934}, 2018.

\bibitem{PNAS}
Chenxi Liu, Barret Zoph, Maxim Neumann, Jonathon Shlens, Wei Hua, Li-Jia Li,
  Li~Fei-Fei, Alan Yuille, Jonathan Huang, and Kevin Murphy.
\newblock Progressive neural architecture search.
\newblock In {\em European Conference on Computer Vision (ECCV)}, pages 19--34,
  2018.

\bibitem{DARTS}
Hanxiao Liu, Karen Simonyan, and Yiming Yang.
\newblock {DARTS}: Differentiable architecture search.
\newblock In {\em International Conference on Learning Representations (ICLR)},
  2019.

\bibitem{IRG}
Yufan Liu, Jiajiong Cao, Bing Li, Chunfeng Yuan, Weiming Hu, Yangxi Li, and
  Yunqiang Duan.
\newblock Knowledge distillation via instance relationship graph.
\newblock In {\em IEEE Conference on Computer Vision and Pattern Recognition
  (CVPR)}, 2019.

\bibitem{TA_distillation}
Seyed-Iman Mirzadeh, Mehrdad Farajtabar, Ang Li, and Hassan Ghasemzadeh.
\newblock Improved knowledge distillation via teacher assistant: Bridging the
  gap between student and teacher.
\newblock {\em arXiv preprint arXiv:1902.03393}, 2019.

\bibitem{parallel-wavenet}
Aaron van~den Oord, Yazhe Li, Igor Babuschkin, Karen Simonyan, Oriol Vinyals,
  Koray Kavukcuoglu, George van~den Driessche, Edward Lockhart, Luis~C Cobo,
  Florian Stimberg, et~al.
\newblock Parallel wavenet: Fast high-fidelity speech synthesis.
\newblock {\em arXiv preprint arXiv:1711.10433}, 2017.

\bibitem{RKD}
Wonpyo Park, Dongju Kim, Yan Lu, and Minsu Cho.
\newblock Relational knowledge distillation.
\newblock In {\em IEEE Conference on Computer Vision and Pattern Recognition
  (CVPR)}, 2019.

\bibitem{penalizing_confidence}
Gabriel Pereyra, George Tucker, Jan Chorowski, {\L}ukasz Kaiser, and Geoffrey
  Hinton.
\newblock Regularizing neural networks by penalizing confident output
  distributions.
\newblock {\em International Conference on Learning Representations (ICLR)},
  2017.

\bibitem{ENAS}
Hieu Pham, Melody Guan, Barret Zoph, Quoc Le, and Jeff Dean.
\newblock Efficient neural architecture search via parameters sharing.
\newblock In {\em International Conference on Machine Learning (ICML)}, pages
  4095--4104, 2018.

\bibitem{FitNets}
Adriana Romero, Nicolas Ballas, Samira~Ebrahimi Kahou, Antoine Chassang, Carlo
  Gatta, and Yoshua Bengio.
\newblock Fitnets: Hints for thin deep nets.
\newblock In {\em International Conference on Learning Representations (ICLR)},
  2015.

\bibitem{ILSVRC15}
Olga Russakovsky, Jia Deng, Hao Su, Jonathan Krause, Sanjeev Satheesh, Sean Ma,
  Zhiheng Huang, Andrej Karpathy, Aditya Khosla, Michael Bernstein,
  Alexander~C. Berg, and Li~Fei-Fei.
\newblock {ImageNet Large Scale Visual Recognition Challenge}.
\newblock {\em International Journal of Computer Vision (IJCV)},
  115(3):211--252, 2015.

\bibitem{CL}
Guocong Song and Wei Chai.
\newblock Collaborative learning for deep neural networks.
\newblock In {\em Advances in Neural Information Processing Systems (NeurIPS)},
  pages 1837--1846, 2018.

\bibitem{DKS}
Dawei Sun, Anbang Yao, Aojun Zhou, and Hao Zhao.
\newblock Deeply-supervised knowledge synergy.
\newblock In {\em IEEE conference on computer vision and pattern recognition
  (CVPR)}, 2019.

\bibitem{ResNeXt}
Saining Xie, Ross Girshick, Piotr Doll{\'a}r, Zhuowen Tu, and Kaiming He.
\newblock Aggregated residual transformations for deep neural networks.
\newblock In {\em IEEE Conference on Computer Vision and Pattern Recognition
  (CVPR)}, 2017.

\bibitem{FSP_matrix}
Junho Yim, Donggyu Joo, Jihoon Bae, and Junmo Kim.
\newblock A gift from knowledge distillation: Fast optimization, network
  minimization and transfer learning.
\newblock In {\em IEEE Conference on Computer Vision and Pattern Recognition
  (CVPR)}, pages 4133--4141, 2017.

\bibitem{metrics_from_teachers}
Lu~Yu, Vacit~Oguz Yazici, Xialei Liu, Joost van~de Weijer, Yongmei Cheng, and
  Arnau Ramisa.
\newblock Learning metrics from teachers: Compact networks for image embedding.
\newblock In {\em IEEE conference on Computer Vision and Pattern Recognition
  (CVPR)}, 2019.

\bibitem{WRN}
Sergey Zagoruyko and Nikos Komodakis.
\newblock Wide residual networks.
\newblock In {\em British Machine Vision Conference (BMVC)}, pages 87.1--87.12.
  BMVA Press, September 2016.

\bibitem{AT}
Sergey Zagoruyko and Nikos Komodakis.
\newblock Paying more attention to attention: Improving the performance of
  convolutional neural networks via attention transfer.
\newblock In {\em International Conference on Learning Representations (ICLR)},
  2017.

\bibitem{ReID}
Xuan Zhang, Hao Luo, Xing Fan, Weilai Xiang, Yixiao Sun, Qiqi Xiao, Wei Jiang,
  Chi Zhang, and Jian Sun.
\newblock Alignedreid: Surpassing human-level performance in person
  re-identification.
\newblock {\em arXiv preprint arXiv:1711.08184}, 2017.

\bibitem{DML}
Ying Zhang, Tao Xiang, Timothy~M Hospedales, and Huchuan Lu.
\newblock Deep mutual learning.
\newblock In {\em IEEE Conference on Computer Vision and Pattern Recognition
  (CVPR)}, 2018.

\bibitem{NAS}
Barret Zoph and Quoc~V. Le.
\newblock Neural architecture search with reinforcement learning.
\newblock In {\em International Conference on Learning Representations (ICLR)},
  2017.

\end{thebibliography}

\section{Appendix}

\subsection{Graph Optimization by ASHA}
\begin{figure}[h]
\centering
\includegraphics[width=0.45\linewidth]{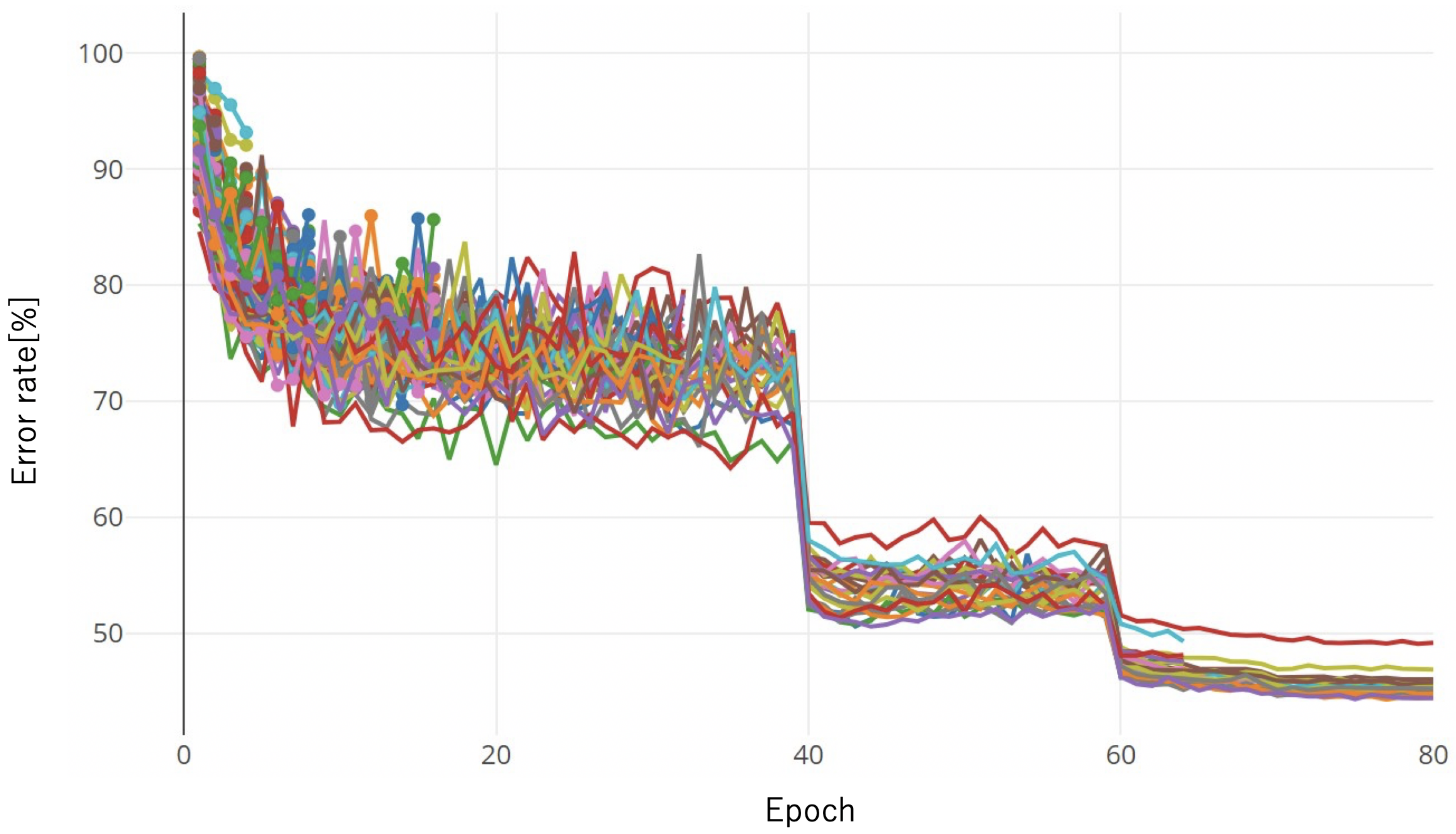}
\caption{\textbf{Error curves when optimizing with ASHA}. We optimized a graph with three nodes on Tiny-ImageNet while pruning with ASHA. Each line represents one trial. The total number of trials is 1500.}
\label{fig:ASHA_optimization}
\end{figure}

Figure~\ref{fig:ASHA_optimization} shows the error curves of each trial in searching a graph using the asynchronous successive halving algorithm~(ASHA)~\cite{ASHA}. ASHA can significantly reduce the computational complexity for searching graphs by terminating trials that are unlikely to achieve a good performance. In our experiments, pruning a trial was performed at the 1, 2, 4, 8, 16, 32, and 64th epochs.

\subsection{Optimization on Reduced Dataset}

\begin{table}[h]
\begin{center}
\begin{tabular}{cc|cc}
Reduction factor & No. of samples & CIFAR-10 & CIFAR-100 \\ \hline
1     & 50000 & 94.14 $\pm$ 0.16 & 74.54 $\pm$ 0.59 \\ 
1/10  & 5000  & 94.08 $\pm$ 0.12 & 74.49 $\pm$ 0.56 \\ 
1/30  & 1666  & 93.83 $\pm$ 0.04 & 74.17 $\pm$ 0.36 \\
1/50  & 1000  & 93.91 $\pm$ 0.19 & 74.29 $\pm$ 0.16 \\
1/70  & 714   & 94.18 $\pm$ 0.12 & 73.67 $\pm$ 0.34 \\
1/100 & 500   & 93.79 $\pm$ 0.12 & 73.11 $\pm$ 0.39 \\
\end{tabular}
\end{center}
\caption{\textbf{Accuracy of target node optimized on reduced dataset}. We optimized graphs with five nodes on the dataset whose number of training samples was reduced and report the average and standard deviation of the recognition rate when the optimized graph was retrained five times using all training data.}
\label{table:reduced_dataset}
\end{table}

We investigated the relationship between the accuracy of a searched top-1 graph and the number of training samples in the dataset used for searching graphs. Table~\ref{table:reduced_dataset} lists the accuracies of graphs optimized on CIFAR-10 and CIFAR-100 with a reduced number of training samples. We report the recognition rate when retraining the top-1 graphs using all training data after searching the graphs. The results show that the performance of graphs tended to decrease when we reduced the number of training samples used for optimization. This suggests that a certain number of samples are required for optimization of the knowledge transfer graph. When the reduction factor is 1/10, reducing the training samples has little impact on the performance of graphs. When the computational cost for searching is also reduced to 1/10, and the search time can be greatly reduced.

\subsection{Visualization of Knowledge Transfer Graphs}
\begin{figure}[h]
%\vspace{-1.5zh}
\centering
\begin{tabular}{ccc}
%--------------------------------3 nodes--------------------------------------
\begin{minipage}[t]{0.32\hsize}
\centering
\includegraphics[width=0.8\linewidth]{03models_each.pdf}
\subcaption{3 nodes, 1st (73.73\%)}
\end{minipage} &
\begin{minipage}[t]{0.32\hsize}
\centering
\includegraphics[width=0.9\linewidth]{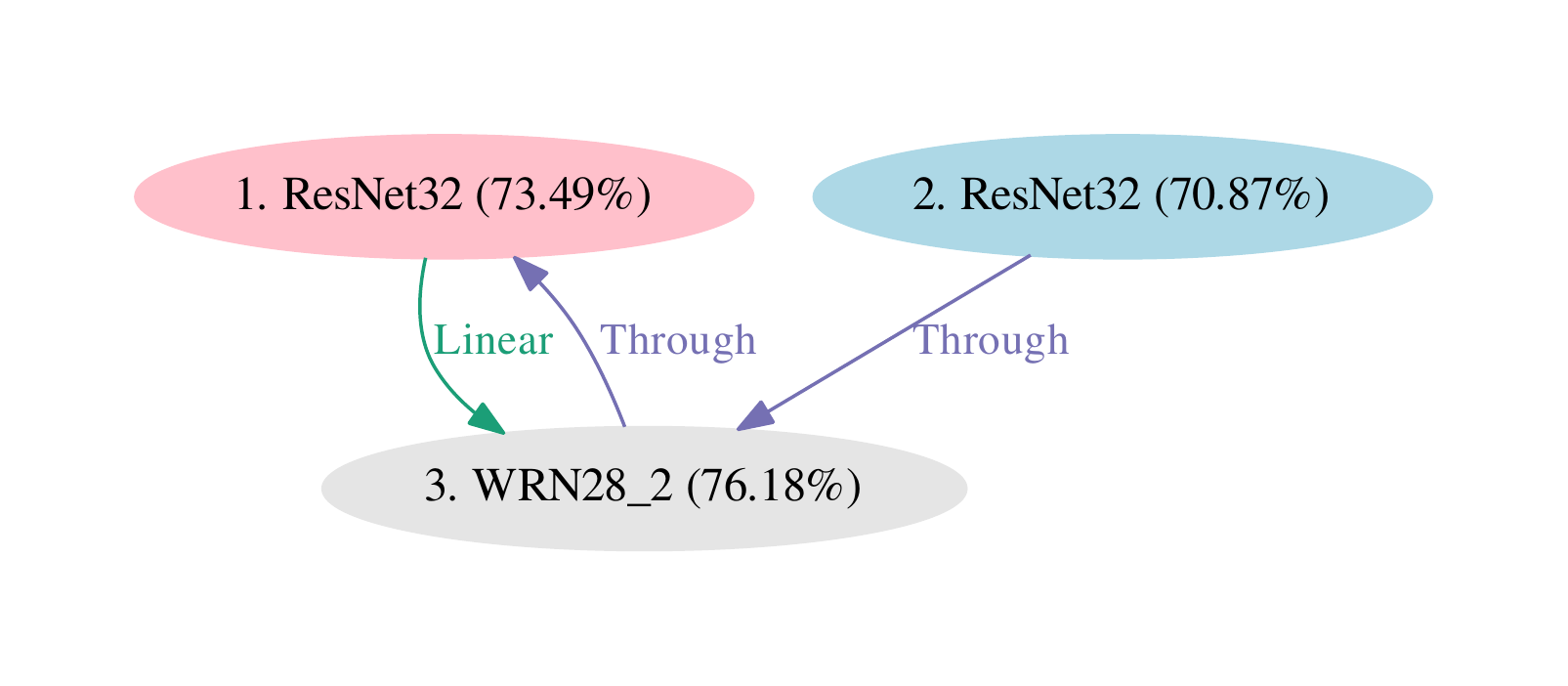}
\subcaption{3 nodes, 2nd (73.49\%)}
\label{fig:top2}
\end{minipage} &
\begin{minipage}[t]{0.32\hsize}
\centering
\includegraphics[width=0.7\linewidth]{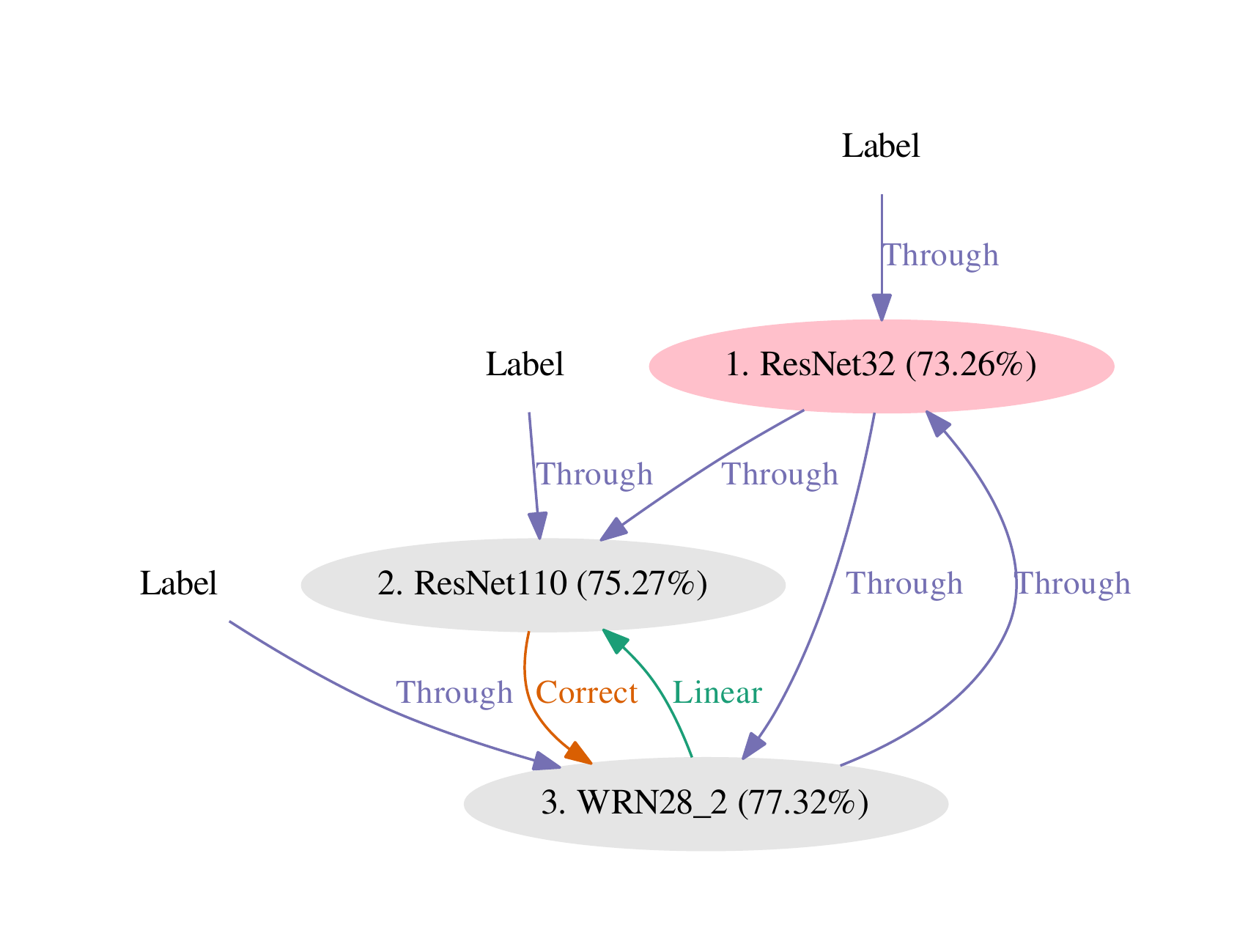}
\subcaption{3 nodes, 3rd (73.26\%)}
\end{minipage} \\
%--------------------------------4 nodes--------------------------------------
\begin{minipage}[t]{0.32\hsize}
\centering
\includegraphics[width=1.0\linewidth]{04models_each.pdf}
\subcaption{4 nodes, 1st (74.60\%)}
\end{minipage} &
\begin{minipage}[t]{0.32\hsize}
\centering
\includegraphics[width=0.8\linewidth]{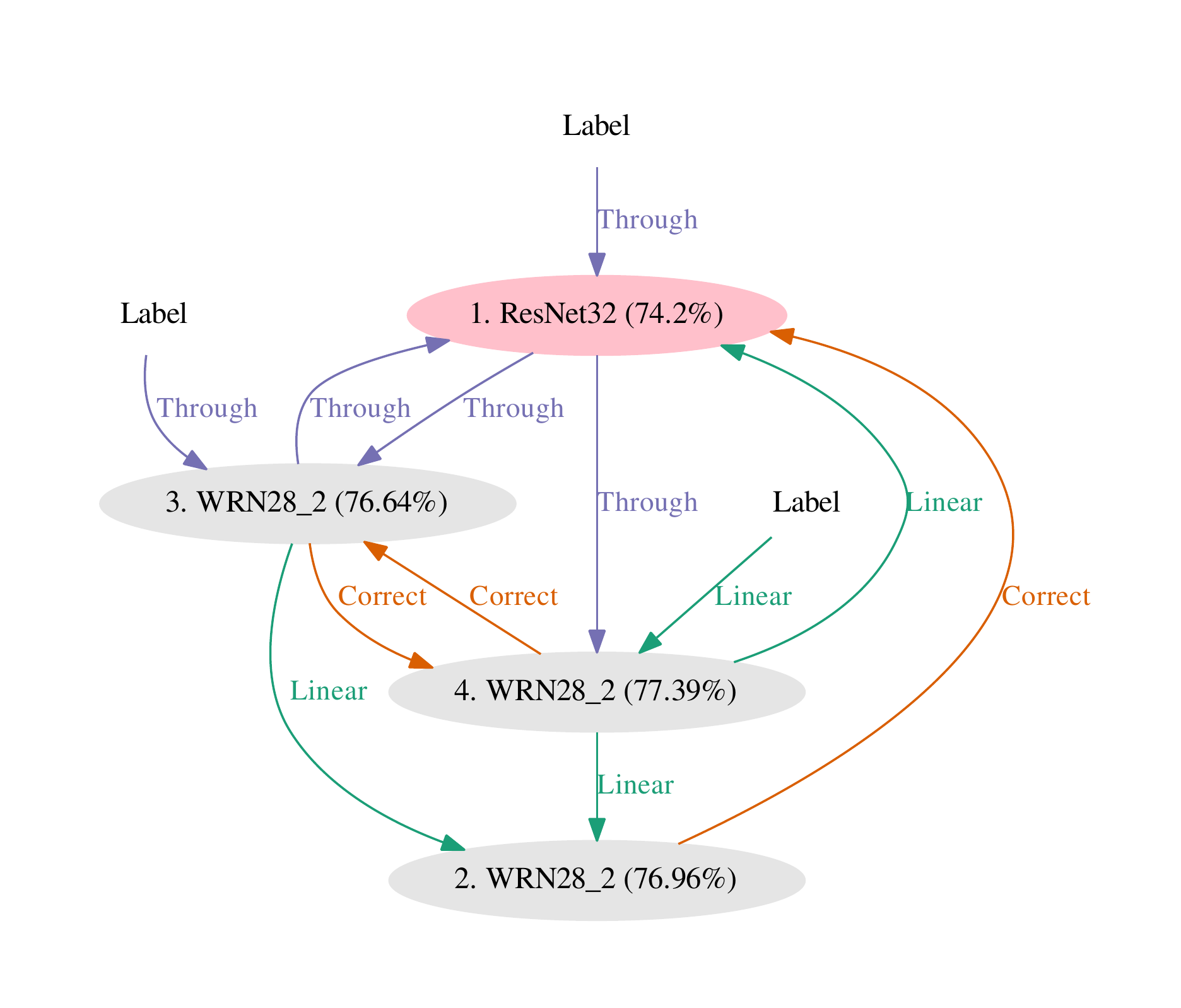}
\subcaption{4 nodes, 2nd (74.20\%)}
\end{minipage} &
\begin{minipage}[t]{0.32\hsize}
\centering
\includegraphics[width=0.8\linewidth]{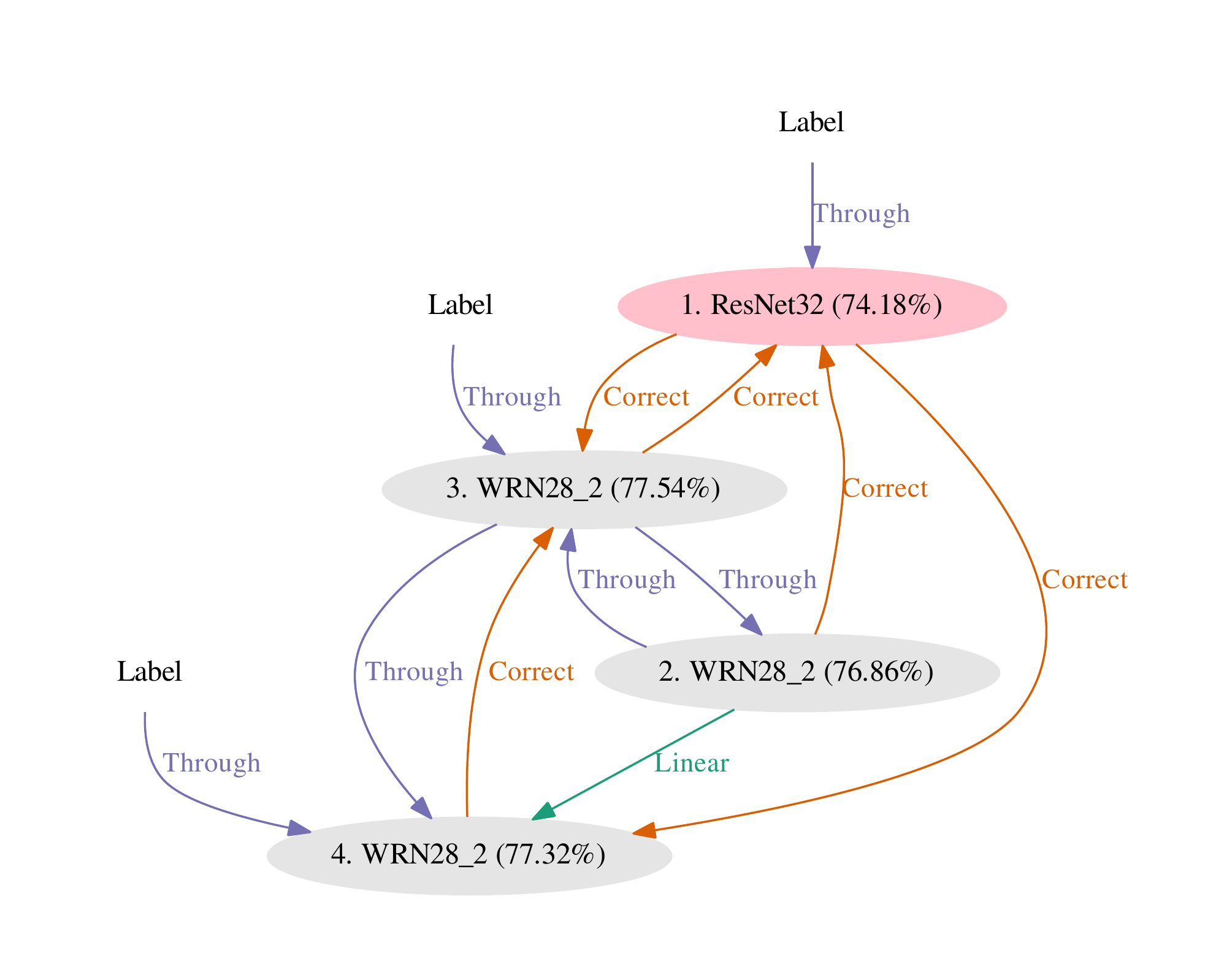}
\subcaption{4 nodes, 3rd (74.18\%)}
\end{minipage} \\
%--------------------------------5 nodes--------------------------------------
\begin{minipage}[t]{0.32\hsize}
\centering
\includegraphics[width=0.7\linewidth]{05models_each.pdf}
\subcaption{5 nodes, 1st (74.51\%)}
\end{minipage} &
\begin{minipage}[t]{0.32\hsize}
\centering
\includegraphics[width=0.9\linewidth]{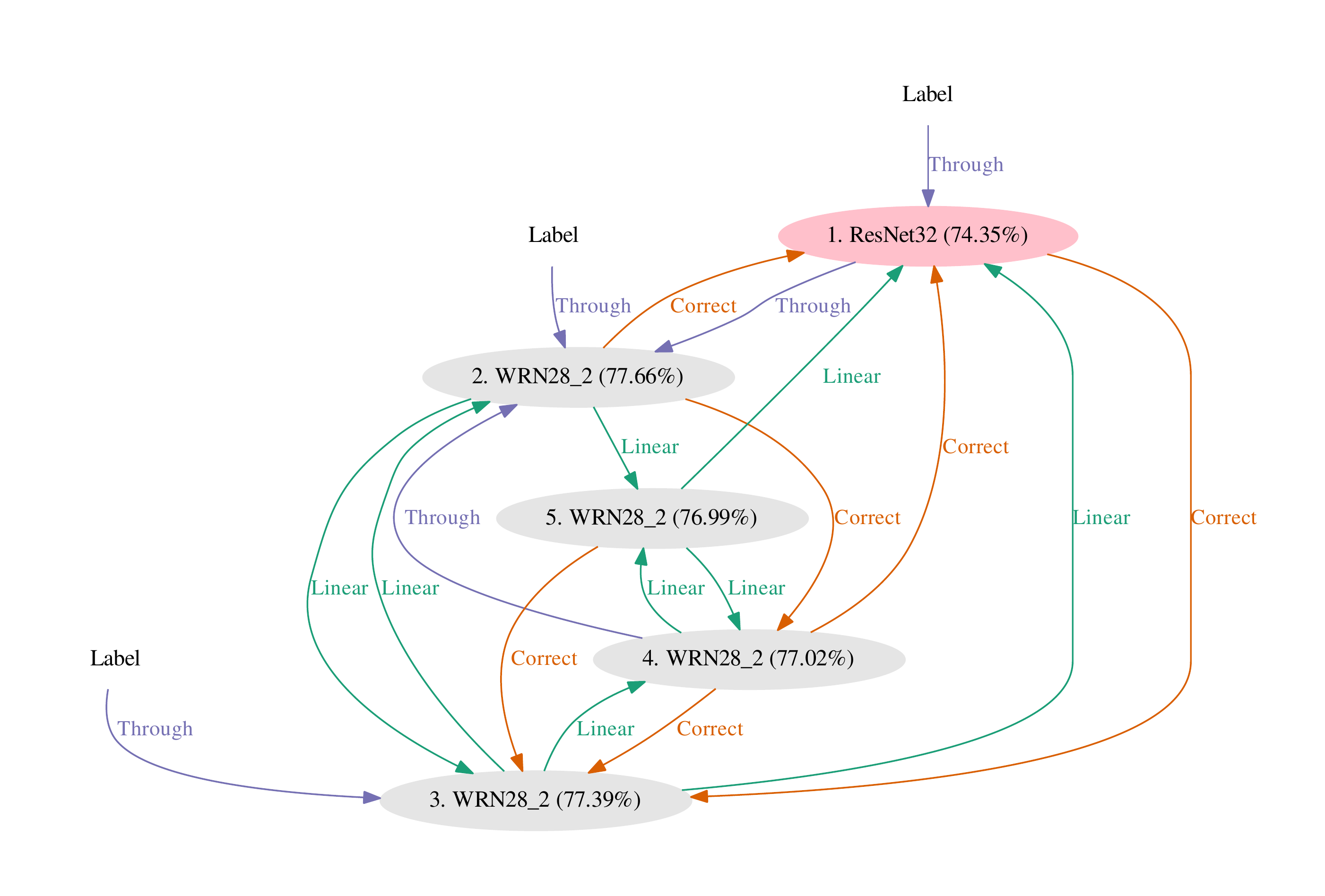}
\subcaption{5 nodes, 2nd (74.35\%)}
\end{minipage} &
\begin{minipage}[t]{0.32\hsize}
\centering
\includegraphics[width=0.9\linewidth]{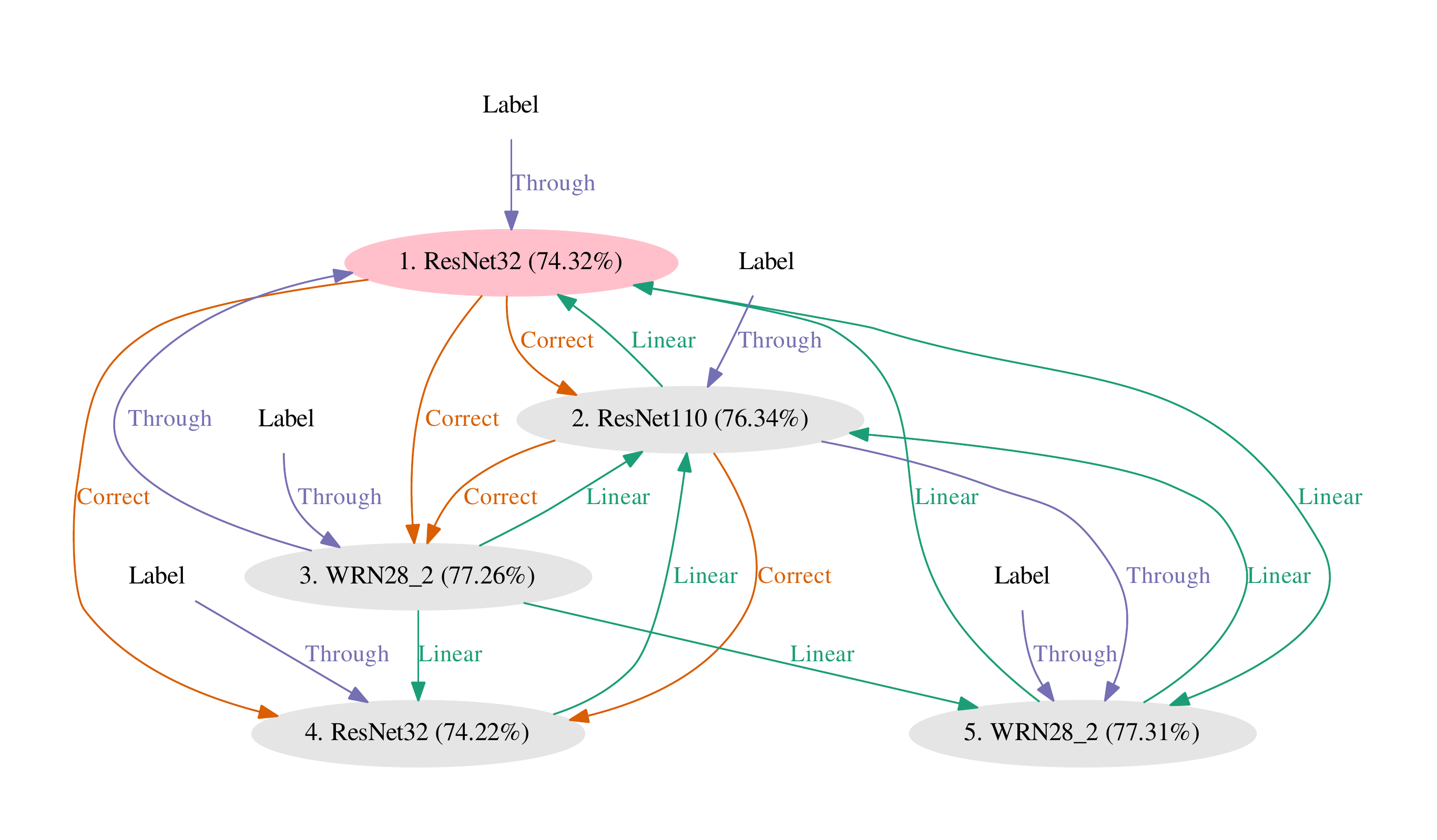}
\subcaption{5 nodes, 3rd (74.32\%)}
\end{minipage} \\
%-------------------------------6 nodes--------------------------------------
\begin{minipage}[t]{0.32\hsize}
\centering
\includegraphics[width=1.0\linewidth]{06models_each.pdf}
\subcaption{6 nodes, 1st (74.46\%)}
\end{minipage} &
\begin{minipage}[t]{0.32\hsize}
\centering
\includegraphics[width=1.0\linewidth]{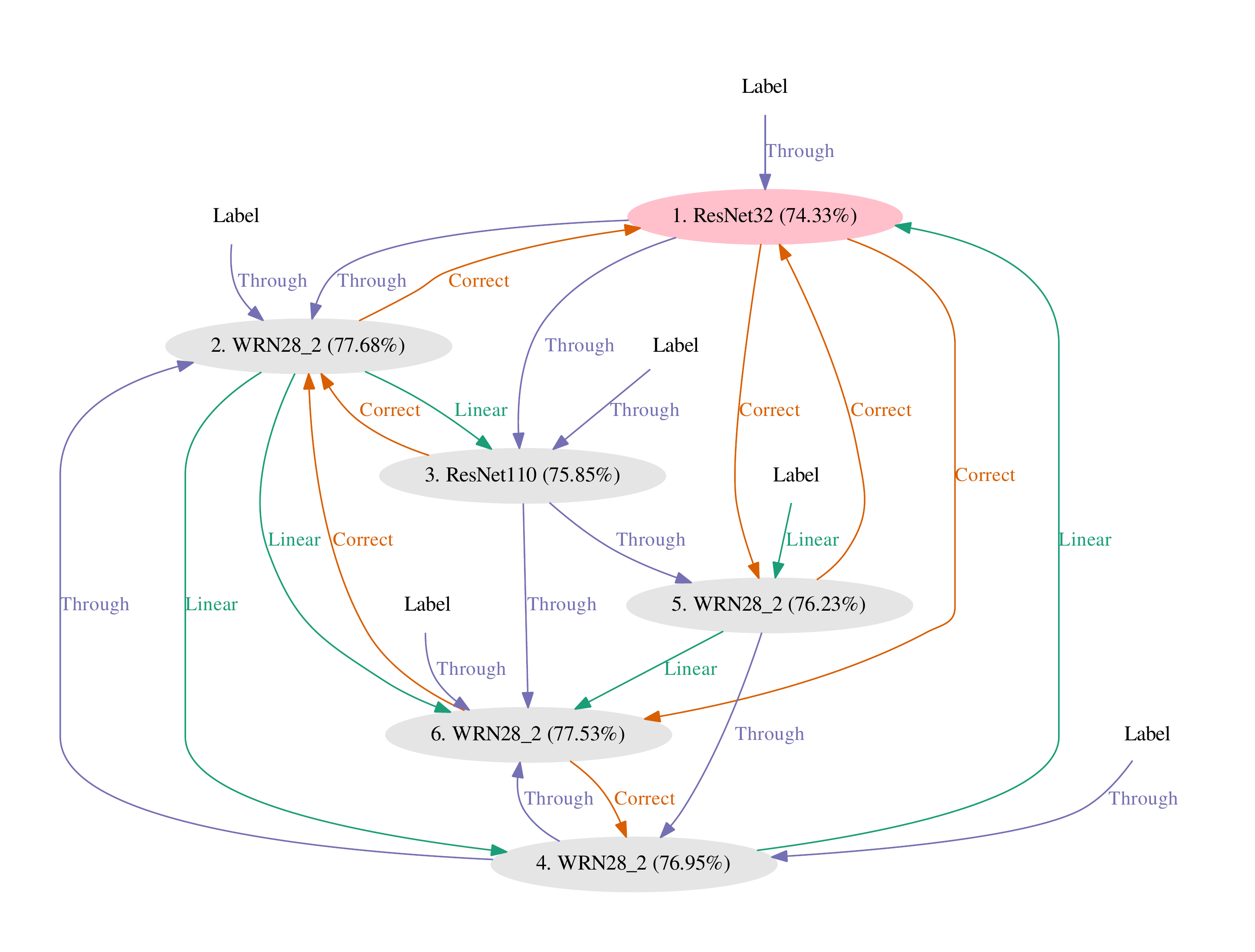}
\subcaption{6 nodes, 2nd (74.33\%)}
\end{minipage} &
\begin{minipage}[t]{0.32\hsize}
\centering
\includegraphics[width=1.0\linewidth]{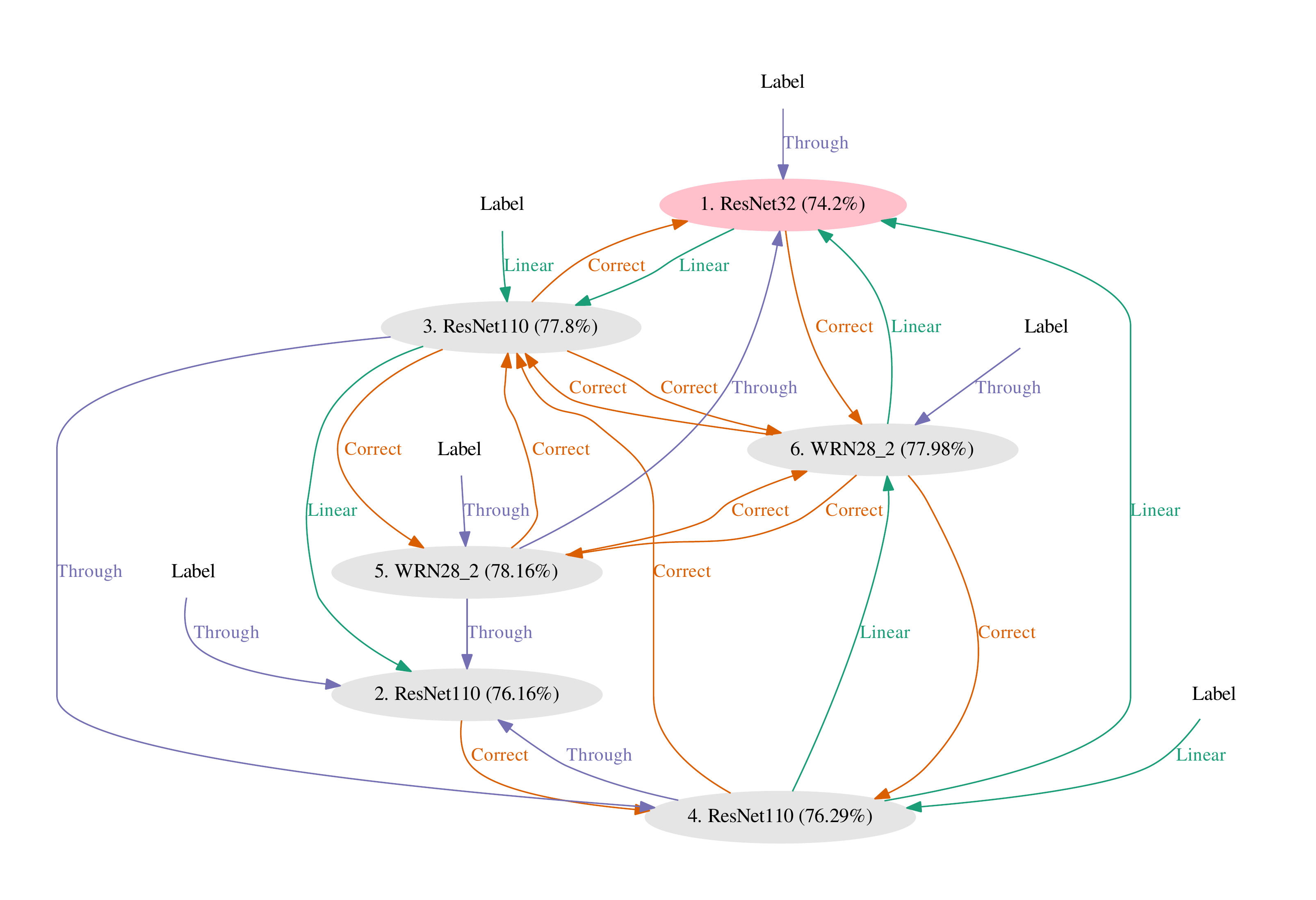}
\subcaption{6 nodes, 3rd (74.20\%)}
\end{minipage} \\
%-------------------------------7 nodes--------------------------------------
\begin{minipage}[t]{0.32\hsize}
\centering
\includegraphics[width=1.1\linewidth]{07models_each.pdf}
\subcaption{7 nodes, 1st (74.82\%)}
\end{minipage} &
\begin{minipage}[t]{0.32\hsize}
\centering
\includegraphics[width=1.1\linewidth]{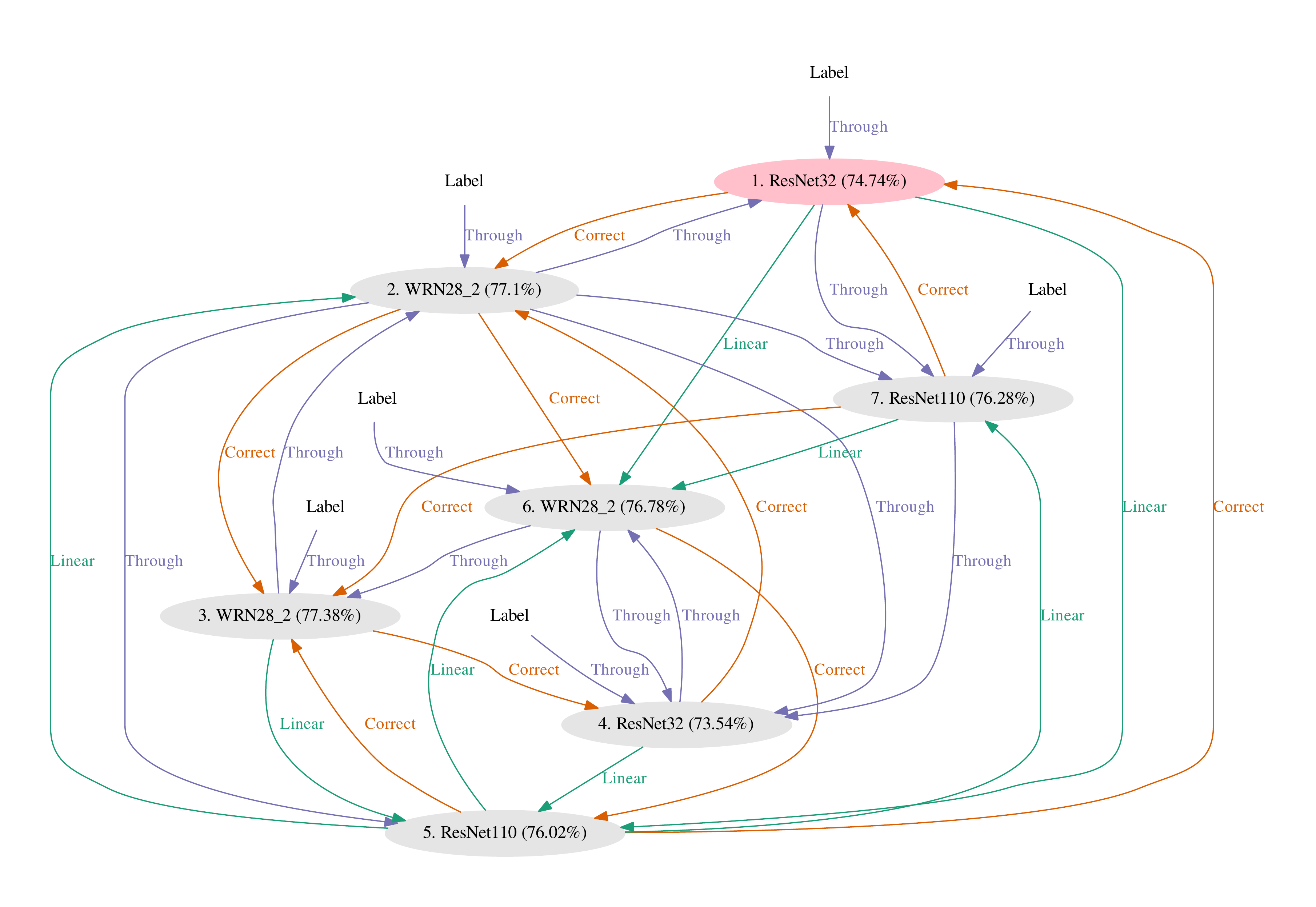}
\subcaption{7 nodes, 2nd (74.74\%)}
\end{minipage} &
\begin{minipage}[t]{0.32\hsize}
\centering
\includegraphics[width=1.1\linewidth]{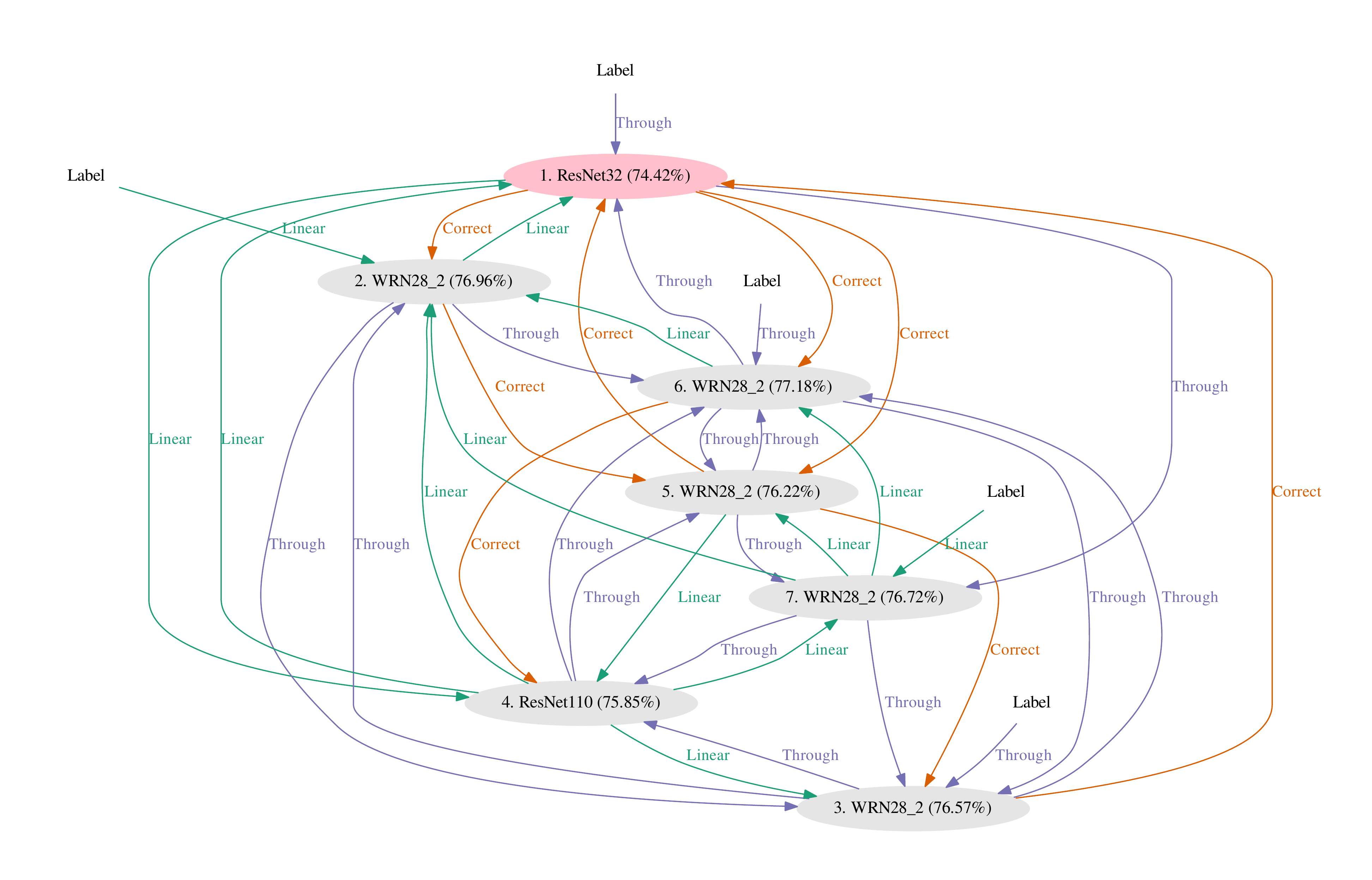}
\subcaption{7 nodes, 3rd (74.42\%)}
\end{minipage} \\

\end{tabular}
\caption{\textbf{Visualization of top-N graphs searched on CIFAR-100}. Red node represents a target node, blue node represents a pre-trained model, and ``Label'' represents a supervised label.}
\label{fig:Top-N_graph}
\end{figure}

Figure~\ref{fig:Top-N_graph} shows the graphs that achieved high performance with three to seven nodes on the CIFAR-100 dataset. The graph shown in Fig.~\ref{fig:top2} is an example of weakly supervised learning. It does not have supervised labels, and nodes 1 and 3 learn from node 2, which is a pre-train model. In addition, this graph can be regarded as self-distillation~\cite{Born_Again_Net} from node 2 to node 1, as these nodes have the same architecture. Node 2 (ResNet32, a small teacher) transfers its knowledge to node 3 (WRN28-2, a large student), and node 3 transfers its knowledge to node 1 (target node). This graph suggests that a network distilled by self-distillation through a larger network can obtain a high generalization ability.

\end{document}